\pdfoutput=1
\documentclass[journal]{IEEEtran}

\IEEEoverridecommandlockouts                              % This command is only needed if 
\usepackage{amsfonts}
\usepackage{mathrsfs}
\usepackage{amsmath}
\usepackage[hidelinks]{hyperref}
\usepackage{url}            % simple URL typesetting
\usepackage{booktabs}       % professional-quality tables
\usepackage{amsfonts}       % blackboard math symbols
\usepackage{lipsum}
\usepackage{graphicx}
\usepackage{svg}
\usepackage{placeins}

\usepackage{soul}
\usepackage{color}

\usepackage{stfloats}
\usepackage{enumitem}

\usepackage{tabularx,booktabs}
\newcolumntype{C}{>{\centering\arraybackslash}X} 
\usepackage{algorithm}
\usepackage{algpseudocode}
\usepackage{amssymb}
\usepackage[most]{tcolorbox}
\usepackage{amsmath,amssymb}

\usepackage{cuted}     % for \begin{strip} ... \end{strip}
\usepackage{capt-of}   % for \captionof{figure} (avoid the 'caption' package with IEEEtran)

\usepackage{float}
\usepackage{mathtools} % Required for rcases command
\title{\LARGE \bf Fast, Robust, Permutation-and-Sign Invariant SO(3) Pattern Alignment}

\author{Anik Sarker$^{1}$ and Alan T. Asbeck$^{1}$% <-this % stops a space
\thanks{*This work was supported by the National Science Foundation, Grant \# 2014499}% <-this % stops a space
\thanks{$^{1}$The authors are with the Dept. of Mechanical Engineering, Virginia Tech, Blacksburg, VA, USA.
        {\tt\small aniks, aasbeck@vt.edu}}
}

\begin{document}

\maketitle
\thispagestyle{empty}
\pagestyle{empty}

%%%%%%%%%%%%%%%%%%%%%%%%%%%%%%%%%%%%%%%%%%%%%%%%%%%%%%%%%%%%%%%%%%%%%%%%%%%%%%%%
%\vspace{-20mm}
% \begin{strip}
% \centering
% \includegraphics[width=\textwidth]{Figures/simulated_dataset.pdf}
% \captionof{figure}{(a) Transfomred Basis vecotrs (TBVs ) represnetation of Example Euroc Mahcihe hal datset MH-05-Diffuicult (A5) as target set. and sumualted set B with a sample ropation RqTBVs and simulated of EuRoC MH-05-Difficult (A5): Example Across Simulated Noise/Outlier Cases (B1-B7)}
% \label{fig:simulated_dataset}
% \vspace{-0.5\baselineskip} % tweak vertical spacing if needed
% \end{strip}

\begin{abstract}
We address the correspondence-free alignment of two rotation sets on \(SO(3)\), a core task in calibration and registration that is often impeded by missing time alignment, outliers, and unknown axis conventions. Our key idea is to decompose each rotation into its \emph{Transformed Basis Vectors} (TBVs)—three unit vectors on \(S^2\)—and align the resulting spherical point sets per axis using fast, robust matchers (SPMC, FRS, and a hybrid). 
To handle axis relabels and sign flips, we introduce a \emph{Permutation-and-Sign Invariant} (PASI) wrapper that enumerates the 24 proper signed permutations, scores them via summed correlations, and fuses the per-axis estimates into a single rotation by projection/Karcher mean. The overall complexity remains linear in the number of rotations (\(\mathcal{O}(n)\)), contrasting with \(\mathcal{O}(N_r^3\log N_r)\) for spherical/\(SO(3)\) correlation. 
% \cite{makadia2006rotation,esteves2023scaling}. 
Experiments on EuRoC Machine Hall simulations 
% \cite{burri2016euroc} 
(axis-consistent) and the ETH Hand-Eye benchmark (\texttt{robot\_arm\_real}) 
% \cite{furrer2017evaluation} 
(axis-ambiguous) show that our methods are accurate, 6-60x faster than traditional methods, and robust under extreme outlier ratios (up to 90\%), all without correspondence search. %This pipeline is a simple, scalable method for \(SO(3)\) set alignment.

\end{abstract}

\section{Introduction}
\label{intro}

Estimating the 3D rotation that aligns one sensor or object frame to another is a fundamental problem in robotics and computer vision. 
Classically posed as \emph{Wahba's problem} \cite{wahba1965least}, it underpins many applications: \emph{point-cloud registration} \cite{besl1992method, chetverikov2002trimmed, li20073d}; \emph{Robot Hand-Eye (RHE) / World-Hand-Eye (RWHE)} pipelines that first estimate rotations, then recover translations, often with time-offset handling or unsynchronized logs \cite{ShiuAhmad1989AXXB,Li2010RWHE,Shah2013Kronecker,Tabb2017RWHE,Furrer2018TimeOffset,furrer2017evaluation}; \emph{IMU-based wearables and gloves} for motion based sensor-to-segment (S2S) calibration, joint-axis identification, and cross-session re-calibration, where orientation distributions alone can recover mounting offsets and axes without time pairing \cite{Seel2012JointAxis,Seel2014IMUAngles,Ekdahl2023S2S,Olsson2020PnP,DiRaimondo2022S2SReview}; \emph{IMU-magnetometer heading calibration}, where yaw alignment is obtained by matching rotation-only statistics while robustly handling hard/soft-iron distortions via ellipsoid-fitting models \cite{Liu2014MagCal,Pang2013MagCal,Wu2020MagCal,Cui2018MagCal}; \emph{VIO/SLAM initialization and relocalization}, where batches of orientations are aligned to refine gravity/yaw and extrinsics before full state estimation \cite{Mourikis2007MSCKF,Qin2018VINSMono,Leutenegger2015OKVIS,Scaramuzza2019VIO}; \emph{molecular alignment} \cite{Thibaut2022}, and more.

Closed-form or least-squares solutions (e.g., Davenport/QUEST, SVD/Procrustes, and modern quaternion solvers) are mature \cite{keat1977analysis,markley1988attitude,markley2000quaternion,wu2017fast,yang2019quaternion}, but they typically assume \emph{paired} measurements (known correspondences) and degrade under heavy outliers or axis-convention mismatches.

%{From spherical correlation to set alignment on \(\mathrm{SO}(3)\).}
A well known strategy treats an \(\mathrm{SO}(3)\) distribution as a spherical signal and aligns two such signals via FFT-based spherical/\(\mathrm{SO}(3)\) cross-correlation \cite{cohen2018spherical}, a technique with roots in 3D shape analysis \cite{kazhdan2003rotation,kazhdan2002harmonic} and used in modern spherical CNNs. 
However, harmonic/correlation pipelines scale as \(\mathcal{O}(N_r^3\log N_r)\) in the rotation-space resolution/bandwidth \cite{sorgi2004normalized,makadia2006rotation}, which can limit practical resolution and scalability \cite{esteves2023scaling}.

In many real scenarios (e.g., multi-sensor rigs, asynchronous logging, or heterogeneous vendors), \emph{pairwise correspondences are unavailable}, sampling rates differ, and axis conventions may be permuted and/or sign-flipped.
A particularly important use case is Robot Hand-Eye / World-Hand-Eye calibration \cite{wise2025certifably, li2010simultaneous, furrer2017evaluation, wang2022accurate, shah2013solving, tabb2017solving, horn2023extrinsic, evangelista2023graph, ulrich2023uncertainty}, which seeks \(X\) (hand\(\rightarrow\)sensor) and \(Y\) (world\(\rightarrow\)robot) satisfying
\[
A_i X \;=\; Y B_i,\quad i=1,\dots,N,\quad A_i,B_i\in SO(3).
\]
Splitting into rotation and translation yields the rotational constraints
\(
R_{A_i}\,X_R \,=\, Y_R\,R_{B_i}
\),
typically solved first on \(SO(3)\), followed by translations. A common two-stage practice is to ignore nonlinear constraints, solve the overdetermined linear system in least squares, then project the resulting matrices onto \(SO(3)\) \cite{li2010simultaneous}; translations are then recovered trivially given the rotations. In practice, however, most RHE/RWHE pipelines \emph{assume synchronized, paired poses}; when sampling rates differ, authors time-align to build pseudo correspondences \cite{furrer2017evaluation, wang2022accurate, shah2013solving, tabb2017solving, horn2023extrinsic, evangelista2023graph, ulrich2023uncertainty}.

\emph{In this work we target the rotational subproblem only.} We estimate, from two \emph{unpaired} rotation sets \(\mathcal{A}=\{R_{A_i}\}\) and \(\mathcal{B}=\{R_{B_j}\}\), a single \(R^\star\in SO(3)\) that best aligns \(\mathcal{B}\) to \(\mathcal{A}\) without time alignment or correspondences. This correspondence-free \(SO(3)\) alignment serves as a robust initializer or intermediate relative-frame estimate for modern (possibly certifiable) RWHE solvers \cite{wise2025certifably}, while leaving translation estimation outside the scope of this paper.

%\vspace{-10mm}
\begin{figure*}
\centering
\includegraphics[width=\textwidth]{Figures/All_steps.pdf}
% \captionof{figure}
\caption{The figure shows how to align two sets of ($SO_3$) distribution using SO3\_SPMC. The major steps of the SO3\_SPMC alignment algorithm in axis consistent setting are depicted through subfigures (a-h). Sub-figures show the progression from one step to another.\\
(a) Two sets of orientations ($SO_3$) are presented in two separate blocks. (b) S2-distributions of two sets and corresponding mean directions are presented. The darker colors represent the $S2^{A}_{x}, S2^{A}_{y}, S2^{A}_{z}$, while the lighter colors correspond to $S2^{B}_{x}, S2^{B}_{y}, S2^{B}_{z}$. (c) In the three-sphere visualization, pairs of $(S2^{AN}_{x},S2^{BN}_{x})$, $(S2^{AN}_{y},S2^{BN}_{y})$, and $(S2^{AN}_{z},S2^{BN}_{z})$ are depicted. (d) The projected S2-distributions are visualized using a 2D histogram. The color legends in the histogram maintain the same notation as mentioned earlier, where darker colors represent the relevant 2D histogram of set $\mathcal{A}$, while lighter colors correspond to set $\mathcal{B}$. (e) After performing the 1D cross-correlation, the respective 2D histograms are aligned. (f) The 2D histogram data, represented in geographic coordinates, is converted to 3D Cartesian coordinates. (g) All S2-distributions are rotated by corresponding inverse of the $R^{A}_{xN}, R^{A}_{zN}, R^{A}_{zN}$. (h) Block diagram illustrates the process of extracting an set of $SO_3$ from the aligned $S2_x$, $S2_y$, and $S2_z$-distributions of set $\mathcal{B}$.}
\label{fig:all_steps}
% \vspace{-0.5\baselineskip} % tweak vertical spacing if needed
\end{figure*}

We recast \(SO(3)\) set alignment as \emph{three} correspondence-free alignments on \(S^2\): each rotation is decomposed into \emph{Transformed Basis Vectors} (TBVs), yielding three spherical point sets per distribution. We then apply fast, robust spherical matchers from prior work \cite{sarker2025correspondence}—SPMC, FRS, and their hybrid—to each axis and \emph{reconstruct} a single \(3\times3\) rotation by projection onto \(SO(3)\). To handle unknown axis conventions, we make the pipeline \emph{Permutation-and-Sign Invariant (PASI)} by exhaustively scoring signed axis permutations and selecting the maximizer (restricting to right-handed cases when appropriate). Fig.~\ref{fig:all_steps} provides an overview of one variant of our method.

\subsection{Contributions}\label{sec:intro_contrib}
\textbf{Correspondence-free \(\mathbf{SO(3)}\) formulation with PASI.}
 We align rotation sets via TBVs on \(S^2\) and enforce invariance to axis relabels/sign flips by enumerating signed permutations \(L=PS\) and selecting the hypothesis that maximizes a correlation objective \cite{sarker2025correspondence}.

\textbf{Linear-time algorithms with strong robustness.}
  Each spherical matcher (SPMC, FRS, hybrid) runs in \(\mathcal{O}(n)\) time \cite{sarker2025correspondence}; our \(SO(3)\) variants require a constant number of calls (3 for axis-consistent; 18 precomputed for PASI), thus remaining \(\mathcal{O}(n)\), versus \(\mathcal{O}(N_r^3\log N_r)\) for spherical/\(SO(3)\) correlation \cite{sorgi2004normalized,makadia2006rotation,esteves2023scaling}. Empirically, the pipeline remains accurate under very high outlier rates (up to \(90\%\)) \cite{sarker2025correspondence}.

\textbf{Extensive evaluation on synthetic and real data.}
  We validate axis-consistent alignment on EuRoC MAV Machine Hall sequences \cite{burri2016euroc} and axis-ambiguous alignment on the ETH Hand-Eye benchmark (\texttt{robot\_arm\_real}) \cite{furrer2017evaluation}, solving the \emph{rotational} RWHE component directly on \(SO(3)\) and comparing against time-aligned RANSAC variants.

%###############################################

% \section{Mathematical Preliminaries and Concepts}

\section{Mathematical Preliminaries and Concepts}

\subsection{Rotating 3D Cartesian Points}\label{rotate_3d_point_set}
Let the points in \(\mathbf{P}\in\mathbb{R}^{n\times 3}\) and let \(\mathbf{R}\in SO(3)\).
Applying the same rotation to all points uses right-multiplication:
\begin{equation}
\mathbf{P}_{\text{rotated}} \;=\; \mathbf{P}\,\mathbf{R}.
\label{rotate_3d_points}
\end{equation}

\subsection{Rows vs.\ Columns of a Rotation Matrix}
For the active, column-vector convention \(\mathbf{v}'=\mathbf{R}\,\mathbf{v}\) with \(\mathbf{R}\in SO(3)\):
\begin{itemize}
  \item \textbf{Columns:} column \(j\) equals \(\mathbf{R}\,\hat{\mathbf e}_j\); i.e., the rotated canonical axes expressed in the original frame.
  \item \textbf{Rows:} row \(i\) equals \(\hat{\mathbf e}_i^\top\mathbf{R}\), giving components via \((\mathbf{v}')_i=\text{row}_i(\mathbf{R})\cdot \mathbf{v}\); geometrically, original-frame axes expressed in the rotated frame (since \(\mathbf{R}^{-1}=\mathbf{R}^\top\)).
\end{itemize}
Example: a \(z\)-axis rotation by \(\theta\),
\[
R_z(\theta)=
\begin{bmatrix}
\cos\theta & -\sin\theta & 0\\
\sin\theta & \ \cos\theta & 0\\
0&0&1
\end{bmatrix},
\]
has first/second columns as the rotated \(x\)/\(y\) axes, while the first row \([\cos\theta,-\sin\theta,0]\) extracts the \(x\)-component after rotation.

%%%%%%%%%%%%%%%
\subsection{Transformed Basis Vectors (TBVs)} \label{basis_vector}
Let \(R_{A\leftarrow B}\in SO(3)\) map coordinates from frame \(B\) to frame \(A\), i.e., \(v_A = R_{A\leftarrow B} v_B\).
We define \emph{Transformed Basis Vectors (TBVs)} using the rows of \(R_{A\leftarrow B}\):
\[
\mathrm{TBV}_i \;\triangleq\; \mathrm{row}_i\!\left(R_{A\leftarrow B}\right)^\top
= \left(R_{A\leftarrow B}\right)^\top \hat{\mathbf e}^{A}_i,
\quad i\in\{x,y,z\}.
\]
Thus, \(\mathrm{TBV}_x\), \(\mathrm{TBV}_y\), and \(\mathrm{TBV}_z\) are the first, second, and third rows of \(R\) (transposed), respectively; equivalently, they are the columns of \(R^\top\).
Geometrically, \(\mathrm{TBV}_i\) is frame \(A\)'s \(i\)-axis expressed in frame \(B\).
Because \(R\in SO(3)\), the TBVs form an orthonormal triad.
For clarity, the \emph{columns} of \(R_{A\leftarrow B}\) are frame \(B\)'s axes expressed in frame \(A\), and the entries satisfy
\((R_{A\leftarrow B})_{ij}=\hat{\mathbf e}^A_i\!\cdot\!\hat{\mathbf e}^B_j\).
In computation, the row view provides direct component extraction,
\[
(v_A)_i \;=\; \mathrm{row}_i\!\left(R_{A\leftarrow B}\right)\, v_B,
\]
and we adopt this row-centric convention throughout unless stated otherwise.

%%%%%%%%%%%%%%%
\subsection{Set of TBVs From Set of Orientations (\texorpdfstring{$SO_3$}{SO3})} \label{vector_representation}
Given a set of \(n\) orientation samples in \(SO_3\), we store them as an array of rotation matrices of shape \(n \times 3 \times 3\).
From this array, we extract the TBVs, yielding three TBV sets: the sets of \(x\)-TBVs, \(y\)-TBVs, and \(z\)-TBVs.

%%%%%%%%%%%%%%%
\subsection{\texorpdfstring{$S^2$}{S2} Representation and S2-point: Normalized 3D Vector Representation on Sphere Surface}
The \(n\)-sphere framework \((S^n)\), a family of compact manifolds embedded in \(\textrm{R}^{n+1}\), is widely used in computer vision, including viewpoint and surface-normal estimation \cite{henderson1996experiencing, flanders1963differential}.  
On the 2-sphere \((S^2)\), a unit 3D vector is represented by its endpoint on the sphere surface, which we refer to as an ``S2-point'' \cite{LiaoCVPR2019}.  
Thus, a normal vector \(\overrightarrow{OP}\) of unit length corresponds to the point \(P\in S^2\); the coordinates of \(P\) encode the direction of the original unit vector.  
The axes of a global reference frame intersect the sphere at three points (on \(x\), \(y\), \(z\)), which can likewise be viewed as the \(S^2\) representation of that frame.

%%%%%%%%%%%%%%%
\subsection{\texorpdfstring{$S^2$}{S2} Representation of TBVs} \label{S2_dist_from_SO3}
\begin{figure}[t]
  \begin{center}
    \includegraphics[width=0.48\textwidth]{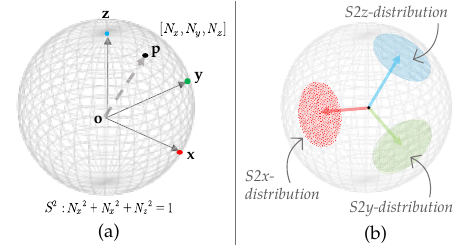}
  \end{center}
 \caption{\(S^2\) representation of a normal vector and \(S^2\) representation of the sets of Transformed Basis Vectors (TBVs).
 (a) In \(S^2\) representation, we view an arbitrary normal vector \(\overrightarrow{OP}\) as a point \(P\) on the surface of the unit sphere; \(P\) is the S2-point of \(\overrightarrow{OP}\).
 Black arrows indicate the reference-frame directions (equivalently, the TBVs of the identity rotation).  
 The red (\(x\)), green (\(y\)), and blue (\(z\)) dots are the S2-points of the identity rotation.
 (b) The \(S2_x\), \(S2_y\), \(S2_z\) distributions are colored red, green, and blue, respectively; their mean directions are shown as arrows.}
\label{fig:vector_point}
\end{figure}

Typically, in the \(n\)-sphere paradigm, a quaternion (in \(\textrm{R}^4\)) is represented on \(S^3\) \cite{LiaoCVPR2019}.  
Here, we instead use the \(S^2\) representation of normal vectors for orientation-derived features.  
A single rotation matrix induces three S2-points via its $x$-, \mbox{$y$-,} and $z$-TBVs.  
In Fig.~\ref{fig:vector_point}(a), the identity rotation yields three S2-points at the coordinate axes.  
For \(n\) \(SO_3\) samples, we first extract three TBV sets (Section~\ref{vector_representation}) and then represent them as three S2-distributions on the sphere surface, denoted \(S2_x\), \(S2_y\), and \(S2_z\); see Fig.~\ref{fig:vector_point}(b).

%%%%%%%%%%%%%%%
\subsection{\texorpdfstring{$S^2$}{S2} Point-Pattern Registration}
Spherical point-pattern registration seeks the optimal rotation that aligns two $S^2$ distributions (target and source). 
Prior work \cite{sarker2025correspondence} introduced three algorithms—\(SPMC\) (Spherical Pattern Matching by Correlation), \(FRS\) (Fast Rotation Search), and a hybrid \(SPMC\_FRS\)—for robust alignment of spherical signals. 
These methods are correspondence-free and robust to moderate noise as well as extremely high outlier ratios. 
Unlike traditional solutions to the Wahba problem \cite{arun1987least, forbes2015linear, horn1987closed, horn1988closed, khoshelham2016closed, markley1988attitude, saunderson2015semidefinite}, algorithms such as \(SPMC\_FRS\) do not require pairwise correspondences between spherical signals. 
Moreover, the source may contain up to \(90\%\) outliers provided the underlying signal remains sufficiently complete relative to the target.
The output is a rotation that aligns the source spherical distribution to the target, which can be applied to 3D points via Eq.~\ref{rotate_3d_points}. Pseudocode for \(SPMC\), \(FRS\), and \(SPMC\_FRS\) is provided in Appendix~\ref{appendix:spherical_matchers}.

%%%%%%%%%%%%%%%
\subsection{Axis-Consistent vs.\ Axis-Ambiguous}
Each \(R\in SO(3)\) yields \(\mathrm{TBV}_x(R)\), \(\mathrm{TBV}_y(R)\), \(\mathrm{TBV}_z(R)\in S^2\).  
A TBV \emph{sign flip} is the antipodal map \(\mathrm{TBV}_i\mapsto -\mathrm{TBV}_i\) (axis preserved, direction reversed). 
In \emph{axis-consistent} settings (e.g., homogeneous sensors sharing a base frame), TBVs correspond one-to-one—\(x\!\leftrightarrow x\), \(y\!\leftrightarrow y\), \(z\!\leftrightarrow z\)—and differ only by a rotation (e.g., two Xsens IMUs on the same rigid body \cite{anik2022}). 
In \emph{axis-ambiguous} settings (e.g., robot world/hand-eye or cross-vendor rigs), axes may be \emph{relabelled} and/or \emph{sign-flipped} (e.g., source \(x\)-TBV matches target \(-y\)-TBV).

%%%%%%%%%%%%%%%
\subsection{Signed Permutations}
Let \(P\in\mathbb{R}^{3\times 3}\) be a permutation matrix (relabels \(\{x,y,z\}\)) and \(S=\mathrm{diag}(s_x,s_y,s_z)\) with \(s_i\in\{\pm1\}\) (independent sign flips). Define
\begin{equation}
L \,\triangleq\, P\,S \,\in\, O(3),
\quad L\in\{-1,0,1\}^{3\times 3},\quad L\,L^\top=I
\label{eq:signed_perm}
\end{equation}
which, by \emph{left} multiplication, reorders/flips the \emph{rows} of any \(3\times 3\) triad (TBVs under our row-vector convention).  
There are \(3!\times 2^3=48\) hypotheses; if a right-handed triad is required, restrict to the \(24\) with \(\det(L)=+1\), noting \(\det(L)=\mathrm{sgn}(P)\,s_x s_y s_z\).

%%%%%%%%%%%%%%%
\subsection{From TBV Pairings to Configurations}
With target TBVs \(\mathrm{TBV}_x(\mathcal{A}),\mathrm{TBV}_y(\mathcal{A}),\mathrm{TBV}_z(\mathcal{A})\) and source TBVs \(\mathrm{TBV}_x(\mathcal{B}),\mathrm{TBV}_y(\mathcal{B}),\mathrm{TBV}_z(\mathcal{B})\), any \(L=PS\) induces a bijection \(\pi\) and signs \((s_x,s_y,s_z)\), yielding
\begin{equation}
\label{eq:pairings_L}
\begin{aligned}
(\mathrm{TBV}_x(\mathcal{A}),\ s_x\,\mathrm{TBV}_{\pi(x)}(\mathcal{B})),\\
(\mathrm{TBV}_y(\mathcal{A}),\ s_y\,\mathrm{TBV}_{\pi(y)}(\mathcal{B})),\\
(\mathrm{TBV}_z(\mathcal{A}),\ s_z\,\mathrm{TBV}_{\pi(z)}(\mathcal{B})).
\end{aligned}
\end{equation}
For each pair we run our spherical pattern matching algorithms (\(SPMC\), \(FRS\), \(SPMC\_FRS\)) to obtain per-axis rotations \((R_x,R_y,R_z)\) and scores \((c_x,c_y,c_z)\), then select
\[
L^\star \;\in\; \arg\max_{L\in\mathcal{L}_{48}} \big(c_x(L)+c_y(L)+c_z(L)\big).
\]

%%%%%%%%%%%%%%%
\subsection{Right-Handed Constraint (Practical Rule)}
Most frames are proper; enforce
\begin{equation}
\begin{split}
\det(L)\;=\;\det(P)\,s_x s_y s_z \;=\; +1 \\
\Longleftrightarrow \quad
s_x s_y s_z \;=\; \det(P),
\label{eq:proper_rule}
\end{split}
\end{equation}
which halves the search to \(24\) configurations.  
\emph{Examples:} (i) Identity \(\pi\): \(\det(P)=+1\Rightarrow s_x s_y s_z=+1\) (even number of flips).
(ii) Swapping \(x\) and \(y\): \(\det(P)=-1\Rightarrow s_x s_y s_z=-1\) (odd number of flips).

%%%%%%%%%%%%%%%
\subsection{Constructing \texorpdfstring{$L$}{L} in Practice}
We build \(L\) from a mapping string (e.g., “Ax\(\to\)\(-\)By, Ay\(\to\)+Bx, Az\(\to\)+Bz”).  
Let rows \(\{1,2,3\}\) correspond to \(\{Ax,Ay,Az\}\) and columns \(\{1,2,3\}\) to \(\{Bx,By,Bz\}\). Then
\begin{equation}
L_{i,j} \;=\;
\begin{cases}
\ +1, & \text{if A-}i \text{ matches } +\text{B-}j,\\
-1, & \text{if A-}i \text{ matches } -\text{B-}j,\\
\ 0, & \text{otherwise},
\end{cases}
\label{eq:L_entries}
\end{equation}
e.g.,
\(
L=\begin{bmatrix}
0 & -1 & 0\\
1 & 0 & 0\\
0 & 0 & 1
\end{bmatrix}
\)
effects \([\,\!-By;\ +Bx;\ +Bz\,]\) when left-multiplying a \(3\times 3\) triad.

%###############################################################

\section{Problem Statement} \label{problem_statement}

Let \(\mathcal{A}=\{\mathbf{a}_i\}_{i=1}^{N}\) and \(\mathcal{B}=\{\mathbf{b}_j\}_{j=1}^{M}\) be two sets of orientations with \(\mathbf{a}_i,\mathbf{b}_j\in SO(3)\).
We posit that a subset of \(\mathcal{B}\) is related to \(\mathcal{A}\) by an unknown global rotation \(\mathbf{R}\in SO(3)\) and a (possibly) unknown signed permutation \( \mathbf{L}\in\mathcal{L}\subset O(3)\) that models axis relabels and sign flips (Section~\ref{eq:signed_perm}); the axis-consistent case corresponds to \(\mathbf{L}=\mathbf{I}\).
Formally, there exist an index set \(\mathcal{J}\subseteq\{1,\ldots,M\}\) and an injective association \(\sigma:\mathcal{J}\to\{1,\ldots,N\}\) such that
\begin{equation}
  \mathbf{b}_j \;=\; \mathbf{L}^\top\,\mathbf{a}_{\sigma(j)}\,\mathbf{R}\,\boldsymbol{\epsilon}_j,
  \qquad j\in\mathcal{J},
  \label{eq:model_axis_ambiguous}
\end{equation}
where \(\boldsymbol{\epsilon}_j\in SO(3)\) models measurement noise/drift with \(\boldsymbol{\epsilon}_j\approx \mathbf{I}\) for inliers; for \(j\notin\mathcal{J}\) (outliers), \(\mathbf{b}_j\) is unconstrained.

Our goal is to estimate \((\mathbf{L},\mathbf{R})\) \emph{without known correspondences} and under potential size mismatch \(M\neq N\).
A robust formulation is
{
\small{
\[
(\mathbf{L}^\star,\mathbf{R}^\star)\!\in
\arg\min_{\substack{\mathbf{L}\in\mathcal{L}\\ \mathbf{R}\in SO(3)}}\,
\min_{\substack{\mathcal{J}\subseteq\{1,\ldots,M\}\\ \sigma:\mathcal{J}\hookrightarrow\{1,\ldots,N\}}}
\sum_{j\in\mathcal{J}} \rho\!\Big(d\big(\mathbf{b}_j, \mathbf{L}^\top\mathbf{a}_{\sigma(j)}\mathbf{R}\big)\Big),
\]
}
}
\noindent
where \(d(\cdot,\cdot)\) is the geodesic distance on \(SO(3)\) and \(\rho\) is a robust loss (e.g., Huber).
In our correspondence-free approach, we do not solve the inner minimization over \((\mathcal{J},\sigma)\) explicitly; instead, we maximize a permutation- and cardinality-invariant correlation objective induced by the TBVs on \(S^2\) (Sections~\ref{S2_dist_from_SO3}-\ref{eq:pairings_L}), enumerating \(\mathbf{L}\) (proper cases when appropriate) and fusing per-axis estimates into a single rotation.

\begin{figure*}[t!]
\centering
\includegraphics[width=\textwidth]{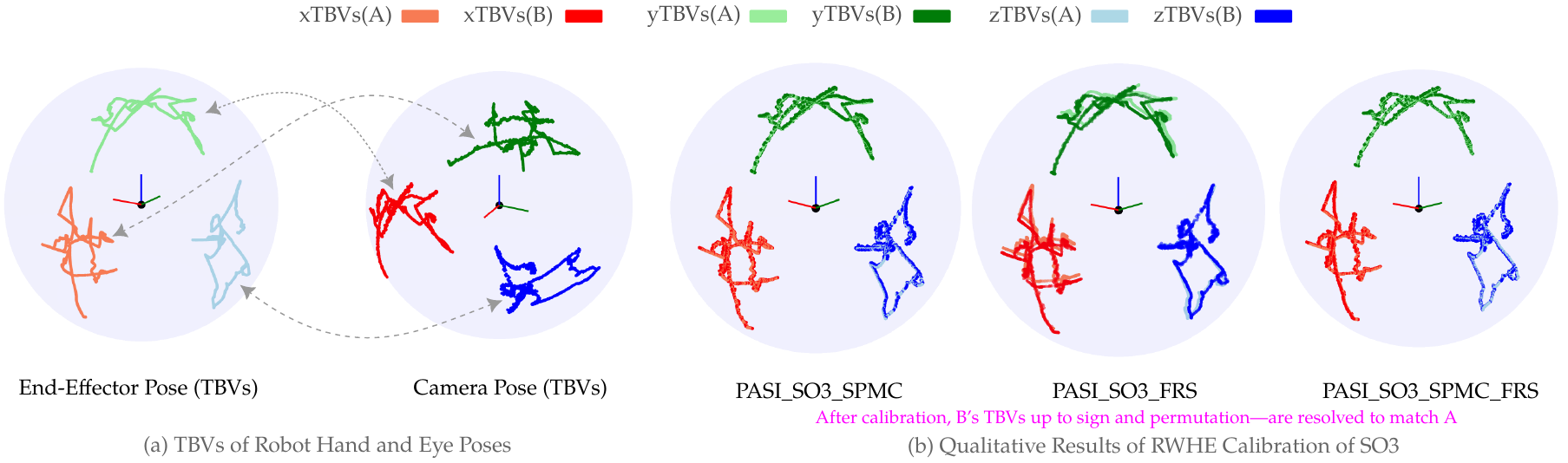}
% \captionof{figure}{
\vspace{-2mm}
  \caption{
\textbf{RWHE on $\mathrm{SO}(3)$ via TBVs and PASI.}
\emph{(a)} Transformed Basis Vectors (TBVs; $S^2$ points) of the end-effector (\(\mathcal{A}\)) and camera (\(\mathcal{B}\)) orientation trajectories from the ETH Hand-Eye Calibration dataset (\texttt{robot\_arm\_real} split) \cite{furrer2017evaluation}. 
The sets exhibit an \emph{axis-ambiguous} relation: the best signed-permutation is
\(L^\star=\bigl[\begin{smallmatrix}0&-1&0\\1&0&0\\0&0&1\end{smallmatrix}\bigr]\),
i.e., \(Ax\!\to\!-By\), \(Ay\!\to\!+Bx\), \(Az\!\to\!+Bz\).
\emph{(b)} Our PASI variant enumerates proper signed permutations \(L=PS\) (\(\det(L)=+1\)), selects \(L^\star\) by maximizing the summed per-axis spherical correlation, then estimates a continuous rotation \(\bar R\) with a spherical matcher (SPMC/FRS/hybrid), yielding the two-sided alignment \(\mathcal{A}\approx L^\star\,\mathcal{B}\,\bar R^\top\).}
\label{fig:rwhec}
\end{figure*}

Figure~\ref{fig:rwhec} provides an example of our method.

%\emph{(c)} \hl{Qualitative overlay and angular-error histogram (computed only for evaluation using time-aligned pairs) show tight post-alignment residuals despite differing set sizes and no correspondence.} \hl{THERE IS NO PART C!!!}

\section{Method: SO(3) Pattern Alignment}
\label{overview_of_sol}

\textbf{Goal.}
Given a \emph{source} orientation set \(\mathcal{B}\subset SO(3)\) and a \emph{target} set \(\mathcal{A}\subset SO(3)\), estimate a single rotation that aligns \(\mathcal{B}\) to \(\mathcal{A}\) so subsequent computations occur in the target frame. In the axis-ambiguous case, we also recover an unknown signed permutation \(L\) that accounts for axis relabels and sign flips (Section II).%(Section~\ref{eq:signed_perm}).

\subsection{From SO(3) to three spherical alignments}
\paragraph{TBVs and \(S^2\) distributions.}
From each rotation we extract row-centric Transformed Basis Vectors (TBVs; Section~\ref{basis_vector}).
Each \(R\in SO(3)\) yields \(\mathrm{TBV}_x(R),\mathrm{TBV}_y(R),\mathrm{TBV}_z(R)\in S^2\), producing three spherical point sets per orientation set: \(S2_x\), \(S2_y\), \(S2_z\) for both \(\mathcal{A}\) and \(\mathcal{B}\) (Section~\ref{S2_dist_from_SO3}).

\paragraph{Per-axis spherical matching.}
Given any TBV pair \((U,V)\subset S^2\), we apply a correspondence-free spherical matcher—\(\texttt{SPMC}\), \(\texttt{FRS}\), or the hybrid \(\texttt{SPMC\_FRS}\) \cite{sarker2025correspondence}—to obtain (i) a rotation \(\widehat R\in SO(3)\) that aligns \(V\) to \(U\), and (ii) a scalar correlation score \(c\). This step is robust to set-size mismatch and outliers and runs in \(\mathcal{O}(n)\) time per call \cite{sarker2025correspondence}. Fig.~\ref{fig:all_steps} visually illustrates all the steps of \(SO3\_SPMC\). Appendix~\ref{appendix:spmc_simulations} provides some numerical simulations of \(SO3\_SPMC\) showing its accuracy and runtime on example synthetic data. 
% \cite{SarkerAsbeck2025SO3Supp}. 

\subsection{Axis-consistent vs.\ PASI pipelines}
\paragraph{Axis-consistent (one-to-one axes).}
When axes match by label and sign (\(x\!\leftrightarrow x\), \(y\!\leftrightarrow y\), \(z\!\leftrightarrow z\)), we align each axis independently:
\[
\begin{aligned}
(R_x,c_x) &= \operatorname{MATCH}(A_x,B_x),\\
(R_y,c_y) &= \operatorname{MATCH}(A_y,B_y),\\
(R_z,c_z) &= \operatorname{MATCH}(A_z,B_z).
\end{aligned}
\]

The three estimates \((R_x,R_y,R_z)\) are fused into a single rotation \(\bar R\in SO(3)\) via a projected arithmetic mean
\(\bar R=\Pi_{SO(3)}\!\big(\tfrac{1}{3}(R_x+R_y+R_z)\big)\) or a Karcher mean. Optionally, a one-shot Procrustes refinement is applied on the reconstructed rotations (Algorithm 1).
%(Section~\ref{algo:all_base}). 

\paragraph{Permutation-and-Sign Invariant (PASI).}
When axes may be relabelled and/or sign-flipped (Section~\ref{S2_dist_from_SO3}), we enumerate signed permutations \(L=PS\) (Eq.~\eqref{eq:signed_perm}). We restrict to the \(24\) proper cases with \(\det(L)=+1\) (Eq.~\eqref{eq:proper_rule}). The PASI procedure is:

\begin{enumerate}[leftmargin=1.1em,itemsep=2pt]
\item \textit{Precompute per-axis matches (18 calls).} For all \(i,j\in\{x,y,z\}\) and \(s\in\{\pm1\}\),
\[
(R^{(s)}_{ij},c^{(s)}_{ij}) \leftarrow \texttt{MATCH}\big(A_i,\ s\,B_j\big).
\]
\item \textit{Score each signed permutation.} Each \(L=PS\) induces the pairings in Eq.~\eqref{eq:pairings_L} via the permutation \(\pi\) and signs \((s_x,s_y,s_z)\).
Score
\[
S(L)=c^{(s_x)}_{x,\pi(x)}+c^{(s_y)}_{y,\pi(y)}+c^{(s_z)}_{z,\pi(z)}.
\]
\item \textit{Select and fuse.} Choose \(L^\star=\arg\max_{L} S(L)\), collect the corresponding \((R_x,R_y,R_z)\), and fuse to a single \(\bar R\in SO(3)\) (projected mean or Karcher mean), followed by optional Procrustes refinement.
\end{enumerate}

Because the 18 per-axis matches are computed once and reused across all \(24\) hypotheses, PASI adds only a lightweight lookup/summation overhead. The three PASI variants—\textbf{PASI\_SO3\_SPMC}, \textbf{PASI\_SO3\_FRS}, \textbf{PASI\_SO3\_SPMC\_FRS}—differ only in the choice of \texttt{MATCH} (SPMC, FRS, SPMC\_FRS \cite{sarker2025correspondence}).

\subsection{Recovered mapping and convention}
\noindent\textbf{Axis-ambiguous alignment (final form).}
Under the row-vector convention, the recovered mapping between sets is
\begin{equation}
\label{eq:two_sided_final}
\mathcal{A} \;\approx\; L^\star\,\mathcal{B}\,\bar R^{\top},
\end{equation}
where \(L^\star\) is the selected proper signed permutation acting on TBV rows, and \(\bar R\in SO(3)\) is the continuous rotation aligning the sign-corrected source to the target. When axes are consistent, \(L^\star=I\) and Eq.~\eqref{eq:two_sided_final} reduces to the axis-consistent case.

\subsection{Algorithm families and baseline structure}
\paragraph{Six instantiations}
All six methods share the same core: (i) extract TBVs; (ii) treat each TBV axis as an \(S^2\) point set; (iii) align via spherical registration; (iv) fuse per-axis rotations.  
\emph{Axis-consistent}: \textbf{SO3\_SPMC} (fastest; may be mean-shift sensitive under heavy outliers), \textbf{SO3\_FRS} (slower; more robust), \textbf{SO3\_SPMC\_FRS} (SPMC proposal + few FRS refinements) \cite{sarker2025correspondence}.  
\emph{PASI}: \textbf{PASI\_SO3\_SPMC}, \textbf{PASI\_SO3\_FRS}, \textbf{PASI\_SO3\_SPMC\_FRS} wrap the axis-consistent core with the signed-permutation enumeration above.

\paragraph{Connection to pseudocode}
Algorithm 1 %~\ref{algo:all_base}
(\emph{Axis-Consistent Baselines on \(SO(3)\)}) provides the common structure used by \textbf{SO3\_SPMC}, \textbf{SO3\_FRS}, and \textbf{SO3\_SPMC\_FRS}; only \texttt{MATCH} changes.  
The PASI counterparts follow the \emph{same} inner steps but prepend the 24-way signed-permutation selection. %; a code-grounded version appears in the \emph{TBV-Exhaustive Alignment on \(\mathrm{SO}(3)\)} box.

\begin{tcolorbox}[float,floatplacement=t,
  width=\columnwidth,
  colback=white!97!gray, colframe=black!60!black,
  title={\bfseries Algorithm 1: Axis-Consistent Baselines on $SO(3)$
    (Common Structure for \texttt{SO3\_SPMC}, \texttt{SO3\_FRS}, \texttt{SO3\_SPMC\_FRS})},
  sharp corners=all, boxrule=0.4pt,
  breakable, enhanced,
  left=0.8ex, right=0.8ex, top=0.6ex, bottom=0.6ex
]
\label{algo:all_base}
%\footnotesize
\small
\textbf{Inputs.}
Target rotations $\{R^A_k\}_{k=1}^n$, Source rotations $\{R^B_\ell\}_{\ell=1}^m$.
A per-axis spherical matcher \texttt{MATCH}$(\cdot,\cdot)$ instantiated as
\texttt{SPMC}, \texttt{FRS}, or \texttt{SPMC\_FRS}.

\textbf{Outputs.}
Estimated global alignment $R_{\mathrm{est}}\in SO(3)$ (maps Source to Target under the row-vector convention).

\textbf{Procedure.}
\begin{enumerate}[leftmargin=1.1em,itemsep=2pt,parsep=0pt,topsep=2pt]
  \item \emph{Extract TBVs per set.}\\
  $(A_x,A_y,A_z)\!\leftarrow\!\texttt{TBVsfromSO3}(\{R^A_k\})$,\;
  $(B_x,B_y,B_z)\!\leftarrow\!\texttt{TBVsfromSO3}(\{R^B_\ell\})$.
  Each $A_i,B_i$ is an $S^2$ point set (row-centric TBVs; Section~\ref{basis_vector}).

  \item \emph{Align each axis via the chosen matcher.}\\
  $(\_,\widetilde B_x)\!\leftarrow\!\texttt{MATCH}(A_x,B_x)$,\;
  $(\_,\widetilde B_y)\!\leftarrow\!\texttt{MATCH}(A_y,B_y)$,\;
  $(\_,\widetilde B_z)\!\leftarrow\!\texttt{MATCH}(A_z,B_z)$.

  \item \emph{Form mean frames.}\\
  $\text{MeanAligned}=\big[\mu(\widetilde B_x);\mu(\widetilde B_y);\mu(\widetilde B_z)\big]$,\;
  $\text{MeanSource}=\big[\mu(B_x);\mu(B_y);\mu(B_z)\big]$,
  where $\mu(\cdot)$ is the Euclidean mean (normalize if desired).

  \item \emph{Estimate global rotation (Procrustes projection).}\\
  $R_{\mathrm{frame}}=\Pi_{SO(3)}\!\big(\text{MeanAligned}^\top\text{MeanSource}\big)$,\;
  $R_{\mathrm{est}}=R_{\mathrm{frame}}^\top$.
\end{enumerate}

\textbf{Instantiation.}
\begin{itemize}[leftmargin=1.1em,itemsep=1pt,topsep=1pt]
  \item \texttt{SO3\_SPMC}: \texttt{MATCH} $\equiv$ \texttt{SPMC} (fastest; can be mean-shift sensitive under heavy outliers).
  \item \texttt{SO3\_FRS}: \texttt{MATCH} $\equiv$ \texttt{FRS} (slower; more robust).
  \item \texttt{SO3\_SPMC\_FRS}: \texttt{MATCH} $\equiv$ \texttt{SPMC\_FRS} (SPMC proposal + few FRS refinements).
\end{itemize}

\textbf{Notes.}
(i) Structure is identical across the three baselines; only \texttt{MATCH} changes.
(ii) To handle axis ambiguity, wrap this baseline with the PASI enumeration (Section~\ref{overview_of_sol}) to obtain \texttt{PASI\_SO3\_\*}.
\end{tcolorbox}

\section{Complexity Analysis}\label{Big-O-compare}

Prior work \cite{sarker2025correspondence} proved that \(\text{SPMC}\), \(\text{FRS}\), and the hybrid \(\text{SPMC\_FRS}\) each run in \(\mathcal{O}(n)\) time, where \(n\) is the number of points involved in the spherical match (proportional to the number of input rotations per axis). By contrast, spherical/\(SO(3)\) cross-correlation methods based on harmonic transforms or rotation-space discretization scale as \(\mathcal{O}(N_r^{3}\log N_r)\), where \(N_r\) denotes either the resolution of the \(\mathrm{SO}(3)\) grid or the retained bandwidth (number of spherical harmonic coefficients) \cite{sorgi2004normalized, makadia2006rotation}; this cubic-log scaling limits practical resolution and thus accuracy at scale \cite{esteves2023scaling}.

\paragraph{Axis-consistent baselines}
Let \(n_A=|\mathcal{A}|\), \(n_B=|\mathcal{B}|\), and \(n\triangleq n_A+n_B\).
Each per-axis spherical match (\(\text{SPMC}\), \(\text{FRS}\), or \(\text{SPMC\_FRS}\)) costs \(\Theta(n)\).
The axis-consistent baselines (\textbf{SO3\_SPMC}, \textbf{SO3\_FRS}, \textbf{SO3\_SPMC\_FRS}) perform three such matches (one for each of \(x,y,z\)):
\[
T_{\text{axis-consistent}} \;=\; 3\,\Theta(n) \;+\; \mathcal{O}(n) \;=\; \mathcal{O}(n),
\]
where the additional \(\mathcal{O}(n)\) term accounts for simple reductions (means) and a constant-time 3\(\times\)3 projection to \(SO(3)\).

\paragraph{PASI variants}
Naïvely, enumerating all \(24\) proper signed permutations and matching three axes per hypothesis would cost \(24\times 3\,\Theta(n)\).
Instead, as detailed in Section~\ref{overview_of_sol}, we \emph{precompute} the \(18\) per-axis matches for all \((i,j,s)\in\{x,y,z\}\times\{x,y,z\}\times\{\pm1\}\), and then evaluate the \(24\) hypotheses by table lookup and summation:
\[
T_{\text{PASI}} \;=\; 18\,\Theta(n) \;+\; \mathcal{O}(24) \;+\; \mathcal{O}(n) \;=\; \mathcal{O}(n).
\]
The final projection (or Karcher mean) and optional Procrustes refinement add only linear accumulation over rotations plus a constant-time 3\(\times\)3 SVD.

\paragraph{Hybrid and refinement costs}
For \textbf{SO3\_SPMC\_FRS}, let \(K\) be the (small) number of FRS refinements; the cost is \(\Theta(n) + K\,\Theta(n) = \Theta(n)\) with a modest constant.  
The Procrustes step forms \(\sum_{\ell}\widehat R^{B,\mathrm{aligned}}_\ell (R^B_\ell)^\top\) in \(\mathcal{O}(n_B)\) and applies a 3\(\times\)3 SVD (constant).

All proposed \(SO(3)\) pattern matchers—\textbf{SO3\_SPMC}, \textbf{SO3\_FRS}, \textbf{SO3\_SPMC\_FRS} and their PASI counterparts—retain \(\mathcal{O}(n)\) time with small constants (3 matches for axis-consistent; 18 precomputed matches for PASI), versus \(\mathcal{O}(N_r^{3}\log N_r)\) for cross-correlation on \(\mathbb{S}^2/\mathrm{SO}(3)\) \cite{sorgi2004normalized, makadia2006rotation, esteves2023scaling}.

\begin{figure*}[t]
  \begin{center}
    \includegraphics[width=0.9\textwidth]{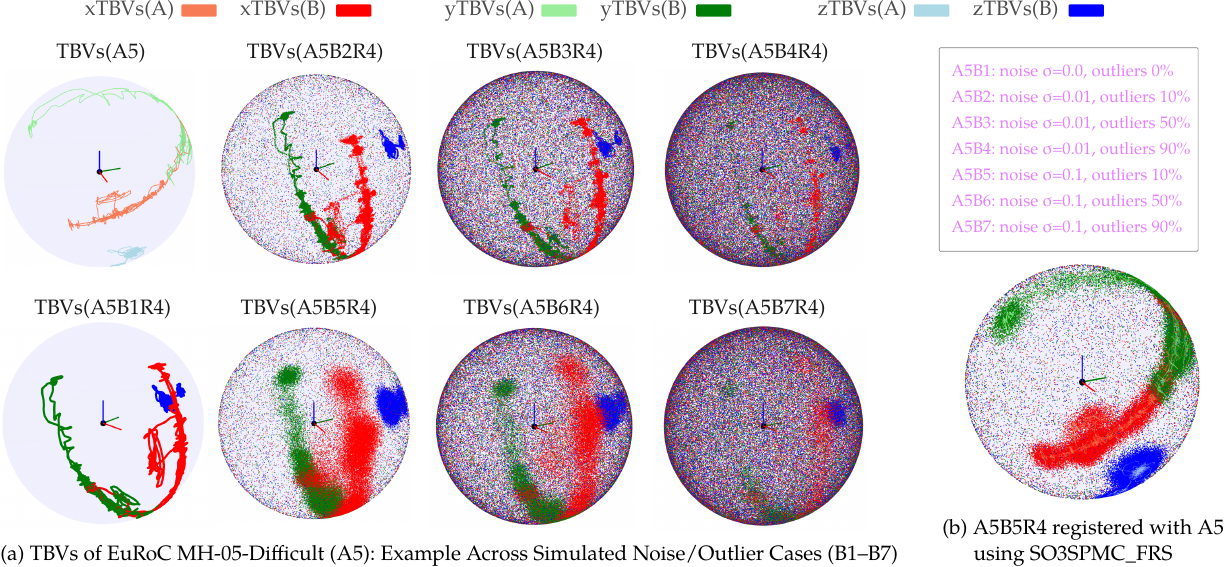}
  \end{center}
  \vspace{-2mm}
  \caption{(a) TBV \(S^2\) distributions for the target set \(A_5\) (EuRoC Machine Hall) and the seven source variants \(B_1\)-\(B_7\) after applying rotation \(R_4\).
  (b) Representative alignment produced by \textbf{SO3\_SPMC\_FRS}.}
  \label{fig:simulated_dataset}
\end{figure*}

\section{Experimental Evaluation}\label{sec:experiments}

We assess both \emph{axis-consistent} and \emph{axis-ambiguous} \(SO(3)\) alignment settings (cf. Section~\ref{overview_of_sol}). 
For the axis-consistent case we run controlled simulations derived from EuRoC MAV Machine Hall sequences \texttt{MH\_01}-\texttt{MH\_05} \cite{burri2016euroc} and test the three axis-consistent aligners. 
For the axis-ambiguous case we solve the \emph{rotational} component of Robot World-Hand-Eye calibration on the ETH Hand-Eye benchmark, split \texttt{robot\_arm\_real} \cite{furrer2017evaluation}, comparing our PASI variants against time-aligned RANSAC baselines (RANSAC Classic (RC) \cite{ransac_original}, RANSAC with pose filtering \cite{schmidt2003robust}, and exhaustive-search (ES) with RANSAC verification over axis configurations).

\subsection{Simulation: Axis-Consistent Scenario}\label{subsec:sim_axis_consistent}
From EuRoC \texttt{MH\_01}-\texttt{MH\_05} we extract orientations to form five target sets \(\mathcal{A}=\{A_1,\dots,A_5\}\) \cite{burri2016euroc}. 
For each \(A_k\), we create seven corrupted source sets \(\{B_1,\dots,B_7\}\).
\begin{itemize}
  \item \textbf{\(B_1\):} no noise, no outliers (same cardinality as \(A_k\)).
  \item \textbf{\(B_2\):} add zero-mean Gaussian noise, std.\ \(0.01\) rad; no outliers.
  \item \textbf{\(B_3\):} keep noise std.\ \(0.01\); replace \(10\%\) of samples with random rotations (outliers).
  \item \textbf{\(B_4\)-\(B_7\):} same noise as \(B_2\); increase outliers to \(25\%\), \(50\%\), \(75\%\), and \(90\%\), respectively.
\end{itemize}
To further stress-test, we sample \(100\) random global rotations \(\{R_1,\dots,R_{100}\}\subset SO(3)\) and apply each to every \(B_j\), yielding \(7\times 100=700\) sources per template \(A_k\) (i.e., \(3500\) trials across \(A_1\)-\(A_5\)). 
We denote configurations as \(\texttt{A5}-\texttt{B5}-\texttt{R4}\): (\(A_5\) is the template; \(B_5\) has \(50\%\) outliers and noise \(0.01\); rotated by \(R_4\)). 
Figure~\ref{fig:simulated_dataset}(a) shows TBVs of a target \(A_k\) and its seven sources under \(R_4\); Fig.~\ref{fig:simulated_dataset}(b) presents a representative registration using \textbf{SO3\_SPMC\_FRS}. 
We report geodesic rotation error, success rates under angle thresholds, and runtime.

\subsection{RWHE Rotational Subproblem: Axis-Ambiguous Scenario}\label{subsec:rwhe_axis_ambiguous}
We adopt the ETH Hand-Eye benchmark (\texttt{robot\_arm\_real}) \cite{furrer2017evaluation}. 
The platform is a UR10 manipulator with an Intel RealSense SR300 RGB-D camera rigidly mounted near the end-effector; an AprilTag target fixed in the environment provides camera poses. 
The dataset logs time-stamped 6-DoF trajectories for the hand in the robot base frame \(T_{B\!H}(t)\) and the eye in the world/target frame \(T_{W\!E}(t)\) as CSVs with fields \(\{t,x,y,z,q_x,q_y,q_z,q_w\}\) (Hamilton convention). 
Here we use \emph{orientation only}: we form two \emph{unpaired} rotation sets,
\(\mathcal{A}=\{R_{B\!H}(t)\}\) (target: end-effector in base) and \(\mathcal{B}=\{R_{W\!E}(t)\}\) (source: camera in world).

Because the frames may differ by axis relabels and sign flips, the axis-consistent methods are not applicable; instead we employ the \emph{PASI} variants: \textbf{PASI\_SO3\_SPMC}, \textbf{PASI\_SO3\_FRS}, and \textbf{PASI\_SO3\_SPMC\_FRS}. 
Each PASI method precomputes the 18 per-axis spherical matches, enumerates the 24 proper signed permutations \(L=PS\) (\(\det(L)=+1\)), scores hypotheses by summed correlations, selects \(L^\star\), and fuses the per-axis rotations into a single \(R^\star\in SO(3)\) aligning \(\mathcal{B}\) to \(\mathcal{A}\). 
We compare against time-aligned RANSAC baselines: RC \cite{ransac_original}, pose-filtered RANSAC \cite{schmidt2003robust}, and ES+RANSAC verification. 
The metrics include geodesic error, success rates, and runtime; translation is out-of-scope, and \(R^\star\) can serve as a robust initializer for full RWHE pipelines.

\section{Results and Discussion}
\label{sec:results}

\begin{figure*}[t]
  \begin{center}
    \includegraphics[width=\textwidth]{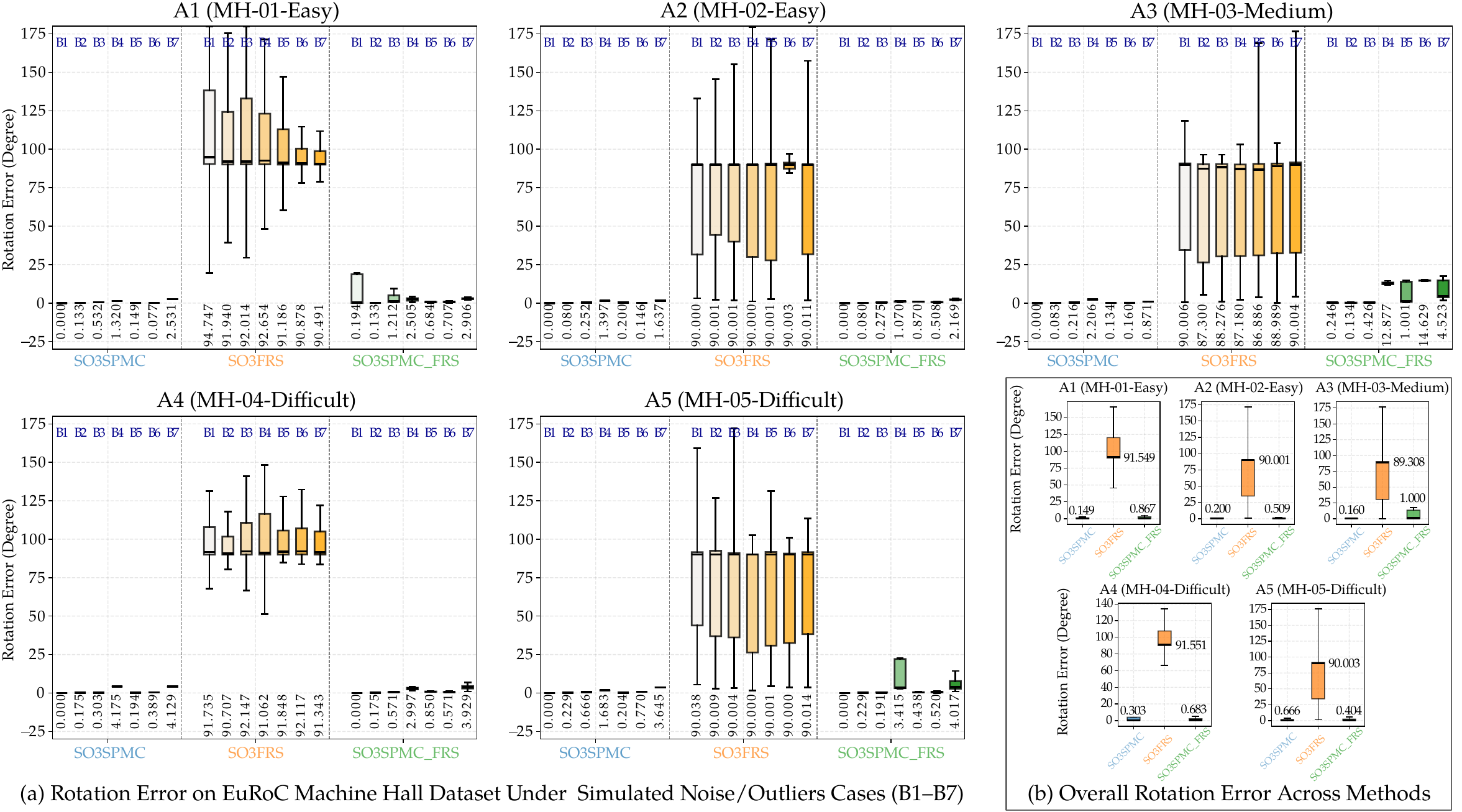}
  \end{center}
  \vspace{-2mm}
  \caption{Axis-Consistent Scenario: Quantitative results on simulated \(SO(3)\) data derived from the EuRoC Machine Hall sequences. 
  For each target set, the seven source configurations \(B1\!\rightarrow\!B7\) (increasing outlier ratio) are shown as box plots; the median of each box is annotated under the plot.}
  \label{fig:euroc_results}
\end{figure*}

\subsection{Simulation: Axis-Consistent Scenario}

Fig.~\ref{fig:euroc_results} summarizes the error distributions across all five EuRoC-derived targets and seven corruption levels.
Three observations are consistent across datasets:

\noindent\textbf{(1) \textit{SO3\_SPMC} is best overall.}
\textbf{SO3\_SPMC} achieves the lowest median geodesic error on the majority of configurations, often with tight interquartile ranges. 
This aligns with prior findings that SPMC excels when the underlying spherical signals are ``complete’’ (i.e., well-overlapped, above $\approx$65\% overlap) \cite{sarker2025correspondence}.

\noindent\textbf{(2) \textit{SO3\_FRS} is consistently suboptimal.}
Across datasets, \textbf{SO3\_FRS} exhibits higher medians and wider tails, reflecting its greater robustness cost when the axis-consistent assumption holds and coverage is good. 
While FRS can be advantageous under heavy contamination, here it underperforms SPMC in accuracy, with the overall rotation error close to 90$^\circ$.

\noindent\textbf{(3) \textit{SO3\_SPMC\_FRS} is competitive but has rare failures.}
The hybrid is generally close to \textbf{SO3\_SPMC} on medians (often \(<\!5^\circ\)), but we observe occasional degradations; e.g., on \texttt{A3}-\texttt{B4} the median reaches \(\sim\!12.87^\circ\).  
A plausible cause is that the FRS refinement, initialized by a good SPMC proposal, can be attracted to a nearby local optimum under certain multimodal or symmetric TBV distributions, leading to a suboptimal final rotation.

Averaging over datasets (Fig.~\ref{fig:euroc_results}b), both \textbf{SO3\_SPMC} and \textbf{SO3\_SPMC\_FRS} perform well, with \textbf{SO3\_SPMC} having the edge in accuracy and stability. \textbf{SO3\_SPMC} has an overall rotation error between 0.15$^\circ$ for the MH-01-Easy set and 0.67$^\circ$ for the MH-05-Difficult set, while the overall rotation error for \textbf{SO3\_SPMC\_FRS} is between 0.40$^\circ$-1.00$^\circ$ for the various sets. 
Our takeaway from this results is when axis correspondences are known (axis-consistent regime), \textbf{SO3\_SPMC} is the recommended default.

%%#######-----------------------------------------------------------

\subsection{Robot World-Hand-Eye on $\mathbf{SO(3)}$: Axis-Ambiguous Scenario}
\label{subsec:rwhe_results}

Fig.~\ref{fig:rwhec}(a) visualizes the TBVs (as \(S^2\) point sets) for the end effector (\(\mathcal{A}\)) and the camera (\(\mathcal{B}\)). 
From the patterns it is evident that while \(z\)-TBVs align across sets, the \(x\)-\(y\) axes are permuted and sign-flipped. 
The selected signed permutation is
\[
L^\star \;=\;
\begin{bmatrix}
0 & -1 & 0\\
1 & \ \,0 & 0\\
0 & \ \,0 & 1
\end{bmatrix},
\]
indicating \(A_x \leftrightarrow -B_y\), \(A_y \leftrightarrow +B_x\), and \(A_z \leftrightarrow +B_z\). 
Hence, axis-consistent baselines are inappropriate here, whereas the PASI variants explicitly handle such relabel/sign ambiguities.

\begin{figure}[t]
  \begin{center}
    \includegraphics[width=0.43\textwidth]{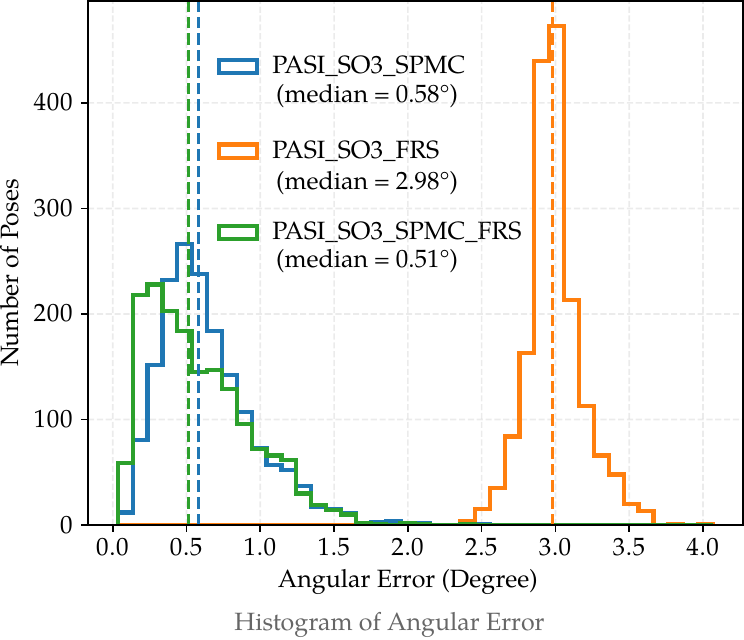}
  \end{center}
  \vspace{-2mm}
  \caption{Axis-Ambiguous Scenario: Histogram of rotational calibration error on the ETH Hand-Eye benchmark (\texttt{robot\_arm\_real}) \cite{furrer2017evaluation}. {SO3\_SPMC} (blue), {SO3\_FRS} (orange), {SO3\_SPMC\_FRS} (green). All algorithms operate on the raw, unpaired rotation streams; for evaluation only, we time-align the trajectories and compute per-pair geodesic angular error on $SO(3)$ (in degrees), enabling a fair comparison to baselines.}
  \label{fig:rwhec_hist}
\end{figure}

Fig.~\ref{fig:rwhec}(b) shows qualitative overlays for \textbf{PASI\_SO3\_SPMC}, \textbf{PASI\_SO3\_FRS}, and \textbf{PASI\_SO3\_SPMC\_FRS}.
Table~\ref{tab:table_rwhec} gives the RMSE and Runtime for each method, while Fig.~\ref{fig:rwhec_hist} reports angular-error histograms. 
For \emph{evaluation only}, we time-align trajectories to form pairs and compute geodesic errors; the \emph{algorithms themselves} consume the raw, unpaired orientation sets (different cardinalities, no synchronization).

Table~\ref{tab:table_rwhec} and Fig.~\ref{fig:rwhec_hist} lead to three conclusions:
(i) \textbf{PASI\_SO3\_SPMC} and \textbf{PASI\_SO3\_SPMC\_FRS} are clearly superior to alternatives in accuracy; 
(ii) our PASI pipeline avoids any correspondence search/synchronization and is \mbox{$\approx6-60\times$} faster than the best time-aligned RANSAC-with-filtering baseline, depending on the variant; 
(iii) among our methods, \textbf{PASI\_SO3\_SPMC} is the fastest (\(\sim 0.7\,\mathrm{s}\)), while \textbf{PASI\_SO3\_SPMC\_FRS} is the most accurate (RMSE \(0.6122^\circ\), median \(0.51^\circ\)). 
\textbf{PASI\_SO3\_FRS} is much less accurate (median and RMSE near \(3^\circ\)).

Overall, PASI delivers correspondence-free rotational RWHE alignment on \(SO(3)\) that is both accurate and efficient in the presence of axis permutations and sign flips; the SPMC-based variant is preferred for speed, and the hybrid for the lowest error.

% \begin{figure*}[t]
%   \begin{center}
%     \includegraphics[width=\textwidth]{Figures/euroc_results.pdf}
%   \end{center}
%   \vspace{-2mm}
%   \caption{Quantitative results on simulated \(SO(3)\) data derived from the EuRoC Machine Hall sequences. 
%   For each target set, the seven source configurations \(B1\!\rightarrow\!B7\) (increasing outlier ratio) are shown as box plots; the median of each box is annotated under the plot.}
%   \label{fig:euroc_results}
% \end{figure*}

\begin{table}[t]
\centering
\setlength{\tabcolsep}{5pt}
\caption{Axis-Ambiguous Scenario: Baseline results on the hand-eye/rwhe task. RMSE reported in degrees; runtime in seconds.}
\label{tab:table_rwhec}
\begin{tabular}{l l r r}
\toprule
\textbf{Method} & \textbf{RMSE ($^\circ$)} & \textbf{Runtime (s)} \\
\midrule

RC\_no\_filter\_best   & 0.6267 & 56.0 \\
RC\_filter\_best       & 0.6677 & 40.7 \\
RS\_filter\_best      & 0.6543 & 44.3 \\
ES\_best               & 0.6543 & 142.0 \\
PASI\_SO3\_SPMC \textbf{(ours)}       & 0.6821 & \textbf{0.7} \\
PASI\_SO3\_FRS \textbf{(ours)}      & 3.0008 & 5.6 \\
PASI\_SO3\_SPMC\_FRS \textbf{(ours)}    & \textbf{0.6122} & 6.2 \\
\bottomrule
\end{tabular}
\end{table}

\section{Conclusion}
We presented a family of correspondence-free alignment algorithms on $SO(3)$ that (i) decompose rotations into \emph{Transformed Basis Vectors} (TBVs), (ii) align the resulting $S^2$ point sets with fast spherical matchers (SPMC, FRS, and a hybrid) \cite{sarker2025correspondence}, and (iii) fuse per-axis estimates into a single rotation. To handle unknown axis conventions, we introduced \emph{Permutation-and-Sign Invariant} (PASI) variants that enumerate proper signed permutations and select the maximizer, making the pipeline invariant to axis relabels and sign flips while preserving efficiency.

The methods accept unpaired sets of differing cardinalities, run in linear time in the number of samples, and remain accurate under high outlier rates. On EuRoC-derived simulations (axis-consistent), \textbf{SO3\_SPMC} is the most accurate and stable default; the hybrid is competitive but can exhibit occasional local-optimum failures on multimodal TBV distributions. On the ETH Hand-Eye benchmark (axis-ambiguous), PASI variants achieve strong accuracy without any time alignment or correspondence search, with only a small constant-factor overhead from enumerating axis configurations.

Beyond benchmarking, our rotation-only initializer is directly useful whenever paired samples are unavailable, including IMU wearables/gloves for sensor-to-segment (S2S) calibration and cross-session re-calibration, IMU-magnetometer heading calibration (with ellipsoid-fit compensation), and VIO/SLAM initialization among others. 
% A particularly promising avenue enabled by TBV+$\mathbb{S}^2$ matching is \emph{automatic orientation calibration} of standalone IMUs: a short \emph{Calibration Task} (CT) produces a \emph{test-time orientation signature}—a session-specific $SO(3)$ distribution—which we align to a \emph{ground-truth orientation signature} (from the same task across prior sessions or from a curated cohort). In axis-consistent deployments (e.g., same device family), our default \textbf{SO3\_SPMC} yields a simple, robust, correspondence-free estimate of the session-wide rotation offset; equivalently, sensor calibration is achieved by aligning the test-time orientation signature with the ground-truth orientation signature. 

\paragraph*{Limitations and future work}
Our pipeline is not yet globally certifiable outside special cases; integrating certifiable $SO(3)$ solvers is a natural direction. Other avenues include streaming/online variants with uncertainty propagation, principled PASI search heuristics, learned activity-conditioned signatures for CT selection, and extensions that couple our rotational initializer with translation recovery for full $SE(3)$ calibration.  We also expect that the overall method could be extended to data with higher dimensions, i.e. $SO(N)$.

Overall, TBV-based alignment with PASI offers a simple, scalable, and robust alternative to spherical/$SO(3)$ correlation, making correspondence-free $SO(3)$ set alignment practical for real-world calibration, registration, and initialization workflows.

% % Name the Tables and Figures differently in the Appendix-- starting with an "A":
% \setcounter{table}{0}
% \renewcommand{\thetable}{A\arabic{table}}
% \renewcommand{\thefigure}{A\arabic{figure}}
% % \renewcommand{\thetable}{\Alph{subsection}\arabic{table}}

\FloatBarrier

% \appendix
\appendices
\section{Spherical Point Pattern Matching Algorithms}
\label{appendix:spherical_matchers}

Following are summaries of the SPMC, FRS, and Hybrid SPMC + FRS algorithms.

%################### SPMC
\begin{tcolorbox}[
  width=\columnwidth,
  colback=white!97!gray, colframe=black!60!black,
  title={\bfseries Algorithm A1: Spherical Pattern Matching by Correlation (SPMC)},
  sharp corners=all, boxrule=0.4pt, breakable, enhanced,
  left=0.8ex, right=0.8ex, top=0.6ex, bottom=0.6ex
]
\footnotesize
\textbf{Input.} Two spherical point sets $\mathcal{A}=\{\mathbf{a}_i\}_{i=1}^N,\ \mathcal{B}=\{\mathbf{b}_j\}_{j=1}^M\subset S^2$. \\
\textbf{Output.} $R_{\text{opt}}\in SO(3)$ aligning $\mathcal{B}$ to $\mathcal{A}$; correlation peak $C(s^*)$.

\textbf{Steps.}
\begin{enumerate}[leftmargin=1.0em,itemsep=2pt,parsep=0pt,topsep=2pt]
\item \emph{Mean-to-North alignment.} Let $\bar{\mathbf{a}}=\tfrac{1}{N}\sum_i\mathbf{a}_i$, $\bar{\mathbf{b}}=\tfrac{1}{M}\sum_j\mathbf{b}_j$. Compute $R_A,R_B\in SO(3)$ such that $R_A\bar{\mathbf{a}}=R_B\bar{\mathbf{b}}=[0,0,1]^\top$.
\item \emph{Rotate sets to the North pole.} $\mathcal{A}^{\mathrm{NP}}=\{ R_A\mathbf{a}_i\}$, $\mathcal{B}^{\mathrm{NP}}=\{ R_B\mathbf{b}_j\}$. (Optional: flip hemisphere so all points have $z\ge 0$.)
\item \emph{Build azimuth histograms.} Form 2D spherical occupancy histograms, then sum over polar bins to get 1D azimuth histograms $h_A,h_B\in\mathbb{R}^{K}$ (e.g., $K\!=\!360$).
\item \emph{1D circular correlation.} For shifts $s\in\{0,\ldots,K-1\}$,
\[
C(s)=\sum_{\lambda=0}^{K-1} h_A(\lambda)\,h_B\!\big((\lambda+s)\bmod K\big).
\]
Let $s^*=\arg\max_s C(s)$.
\item \emph{Recover the rotation.} $R_{\text{shift}}=R_z(2\pi s^*/K)$,\quad $R_{\text{opt}}=R_A^{-1}\,R_{\text{shift}}\,R_B$.
\end{enumerate}

\textbf{Notes.} Exact recovery holds in the noise/outlier-free one-to-one case (means coincide; $z$-axis residual only).
\end{tcolorbox}

% \label{spmc_steps}

%%%### FRS

\begin{tcolorbox}[
  width=\columnwidth,
  colback=white!97!gray, colframe=black!60!black,
  title={\bfseries Algorithm A2: Fast Rotation Search (FRS)},
  sharp corners=all, boxrule=0.4pt, breakable, enhanced,
  left=0.8ex, right=0.8ex, top=0.6ex, bottom=0.6ex
]
\footnotesize
\textbf{Input.} $\mathcal{A}=\{\mathbf{a}_i\}_{i=1}^N,\ \mathcal{B}=\{\mathbf{b}_j\}_{j=1}^M\subset S^2$; max iters $T$; tolerance $\varepsilon$. \\
\textbf{Output.} $R_{\text{est}}\in SO(3)$ aligning $\mathcal{B}$ to $\mathcal{A}$.

\textbf{Steps.}
\begin{enumerate}[leftmargin=1.0em,itemsep=2pt,parsep=0pt,topsep=2pt]
\item \emph{Fix target histograms.} Compute mean-to-North $R_A$; set $\mathcal{A}^{\mathrm{NP}}=\{R_A\mathbf{a}_i\}$. Build fixed azimuth histograms $h^x_A,h^y_A,h^z_A$ (three views / axis-angle parameterizations).
\item \emph{Initialize.} $R\leftarrow I_3$, $t\leftarrow 0$.
\item \emph{Iterate (coarse-to-fine azimuth updates).} While $t<T$:
  \begin{enumerate}[leftmargin=0.9em,itemsep=1pt,parsep=0pt,topsep=0pt]
  \item Form $\mathcal{B}^{(t)}=\{R\,\mathbf{b}_j\}$ and $\mathcal{B}^{\mathrm{NP}}=\{R_A\,\mathcal{B}^{(t)}\}$.
  \item Build moving azimuth histograms $h^x_B,h^y_B,h^z_B$.
  \item For each axis $u\in\{x,y,z\}$, compute circular shift $s_u^*=\arg\max_s \sum_\lambda h^u_A(\lambda)\,h^u_B((\lambda+s)\bmod K)$.
  \item Let $R_{\text{inc}}=R_z(2\pi s_z^*/K)\,R_y(2\pi s_y^*/K)\,R_x(2\pi s_x^*/K)$.
  \item \textbf{Stop} if all $|s_u^*|\le \varepsilon_K$ (equivalently, $\|\log(R_{\text{inc}})\|\le\varepsilon$).
  \item Update $R\leftarrow R_{\text{inc}}\,R$, $t\leftarrow t+1$.
  \end{enumerate}
\item \emph{Return.} $R_{\text{est}}=R$.
\end{enumerate}

\textbf{Notes.} Using SPMC’s mean-to-North initialization and multi-axis azimuth updates yields rapid convergence; small-$T$ (e.g., 5-10) is often sufficient in practice.
\end{tcolorbox}

%%%#######
\begin{tcolorbox}[
  width=\columnwidth,
  colback=white!97!gray, colframe=black!60!black,
  title={\bfseries Algorithm A3: Hybrid SPMC\,+\,FRS},
  sharp corners=all, boxrule=0.4pt, breakable, enhanced,
  left=0.8ex, right=0.8ex, top=0.6ex, bottom=0.6ex
]
\footnotesize
\textbf{Input.} $\mathcal{A},\mathcal{B}\subset S^2$; FRS hyperparameters $(T,\varepsilon)$. \\
\textbf{Output.} $R_{\text{est}}\in SO(3)$.

\textbf{Steps.}
\begin{enumerate}[leftmargin=1.0em,itemsep=2pt,parsep=0pt,topsep=2pt]
\item \emph{SPMC initialization.} Run SPMC (Alg.~A1) to obtain $R_0$.
\item \emph{Warm-start FRS.} Initialize $R\!\leftarrow\!R_0$ and run FRS (Alg.~S2) with small $T$ and tight $\varepsilon$.
\item \emph{Return.} $R_{\text{est}}=R$.
\end{enumerate}

\textbf{Notes.} Combines SPMC’s fast global azimuth recovery with FRS’s local refinement; empirically more robust under moderate outliers than pure SPMC, while remaining faster than pure FRS.
\end{tcolorbox}

\begin{figure*}[t]
  \begin{center}
\includegraphics[width=\textwidth]{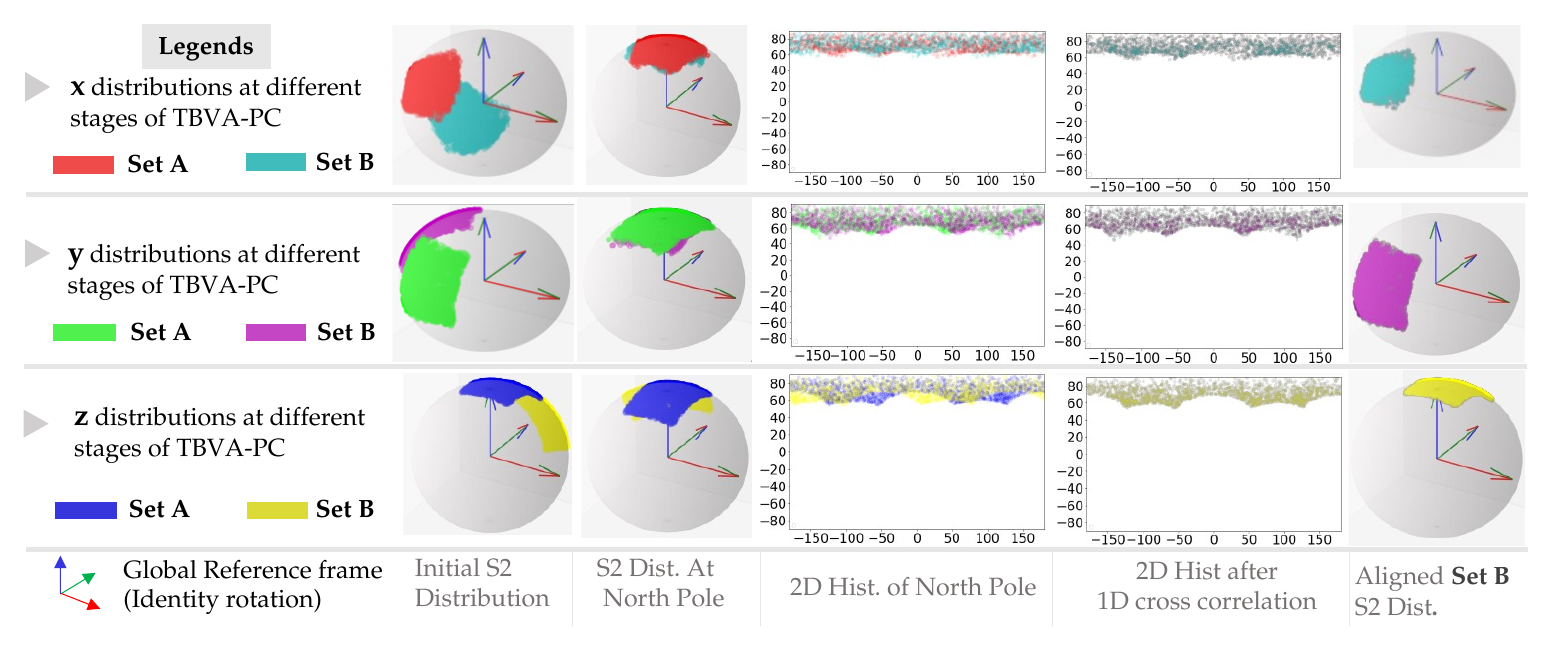}
  \end{center}
  \vspace{-2mm}
 \caption{Visualization of different stages of the SOM\_SPMC algorithm with Scenario~1. The top, middle, and bottom rows correspond to different stages of the x, y, and z distributions respectively. The left-most column presents the initial S2-distributions, while the right-most column demonstrates the final alignment. From the plots in the last column, it is evident that all distributions overlap significantly, indicating a successful alignment.} 
\label{fig:all_steps_visualization}
\end{figure*}

\section{Numerical Simulations of \texorpdfstring{$SO_3$}{SO3} Alignment}
\label{appendix:spmc_simulations}

While each of these algorithms was previously presented and validated in \cite{sarker2025correspondence}, we provide additional examples and validation of the \emph{\texttt{SO3\_SPMC}} algorithm on several commonly occurring spherical patterns. 
\subsection{Numerical Simulations}

A truly uniform sampling of $SO(3)$ induces (approximately) uniform TBV distributions on $S^2$, yielding little discriminative structure. 
To obtain informative test cases, we generate structured $SO(3)$ samples by restricting Euler angles to bounded ranges.

\paragraph{Data generation (set $\mathcal{A}$).}
We sample $n=2000$ Euler-angle triples (roll, pitch, yaw) within the ranges
\[
\text{roll}\in(-40^\circ,10^\circ),\:
\text{pitch}\in(-20^\circ,20^\circ),\:
\text{yaw}\in(-10^\circ,50^\circ).
\]
Each triple is converted to a rotation matrix using standard formulas \cite{slabaugh1999computing}, producing $\mathcal{A}=\{A_i\}_{i=1}^{n}\subset SO(3)$. 
Figure~\ref{fig:all_steps_visualization} illustrates a representative configuration and its $S^2$ (TBV) views.

\paragraph{Ground-truth transform and source set.}
We draw an arbitrary ground-truth $R_{\mathrm{gt}}\in SO(3)$ and synthesize
\[
\mathcal{B}\;=\;\{\,B_i\equiv R_{\mathrm{gt}}\,A_i\,\}_{i=1}^{n}.
\]
We then run \emph{\texttt{SO3\_SPMC}} to align the spherical patterns and recover $R_{\mathrm{est}}$, the estimate of $R_{\mathrm{gt}}$.

\paragraph{Error metric.}
Given $R_1,R_2\in SO(3)$, the geodesic (angular) error is
\[
\theta(R_1,R_2)\;=\;\cos^{-1}\!\Big(\tfrac{\mathrm{trace}(R_1 R_2^\top)-1}{2}\Big)\;[\mathrm{rad}],
\]
converted to degrees for reporting. 
For the synthesized pair $(\mathcal{A},\mathcal{B})$ we form the constant relative error
\[
\Delta \;\equiv\; R_{\mathrm{gt}}\,R_{\mathrm{est}}^\top,
\]
which equals $B_i\,(R_{\mathrm{est}}A_i)^\top$ for each $i$ (since $A_iA_i^\top=I$). 
We report the mean angular error (MAE) $\theta(\Delta)$; for completeness we also compute the per-pair angle and average (which coincides with $\theta(\Delta)$ in this noiseless construction).

\paragraph{Scenarios and sizes.}
We evaluate three pattern families, each at two set sizes ($n=2000$ and $n=10$) to probe both data-rich and data-scarce regimes:
\begin{itemize}
\item \textbf{Scenario 1 (bounded integer ranges).} 
Random Euler triples with integer end-points:
$(-40^\circ,10^\circ)$, $(-20^\circ,20^\circ)$, $(-10^\circ,50^\circ)$.

\item \textbf{Scenario 2 (bounded non-integer ranges).} 
Random Euler triples with non-integer end-points:
$(-40.5^\circ,10.5^\circ)$, $(-20.5^\circ,20.5^\circ)$, $(-10.5^\circ,50.5^\circ)$.
This stresses potential interactions with histogram binning in \texttt{SO3\_SPMC}.

\item \textbf{Scenario 3 (Gaussian around a mean).} 
Quaternions sampled from a Gaussian in angle space, centered at Euler means $[30^\circ,10^\circ,15.5^\circ]$ (roll, pitch, yaw), then mapped to $SO(3)$.
\end{itemize}

\paragraph{Protocol and reporting.}
For each scenario and size, we run \texttt{SO3\_SPMC} to estimate $R_{\mathrm{est}}$, then report MAE in degrees. 
A summary across all settings appears in Table~\ref{Table:simulation_combination}. 
Representative $S^2$ visualizations of the TBV distributions and an end-to-end example are provided in Figure~\ref{fig:all_steps_visualization}.

\begin{table}[ht]
    \centering
    \caption{Mean Angular Error (MAE, degrees) for \texttt{SO3\_SPMC} under various simulation scenarios.}
    \vspace{2pt}
    \begin{tabular}{|c|c|c|}
        \hline
        \textbf{Scenario} & \textbf{2000 points} & \textbf{10 points} \\
        \hline
        1 (bounded integer ranges) & $0.060^\circ$ & $0.10^\circ$ \\
        2 (bounded non-integer ranges) & $0.068^\circ$ & $0.15^\circ$ \\
        3 (Gaussian around a mean) & $0.068^\circ$ & $0.08^\circ$ \\
        \hline
    \end{tabular}
    \label{Table:simulation_combination}
\end{table}

\subsection{Ablation on Number of Points}
\label{supp:ablation_points}

We present how runtime and accuracy scale with the number of rotations. 
For each scenario, we form sets with cardinalities
\[
n\in\{10^1,\,10^2,\,10^3,\,10^4,\,10^5,\,10^6\}
\]
and run \texttt{SO3\_SPMC} to estimate $R_{\mathrm{est}}$.

\begin{figure}[ht]
  \centering
  \includegraphics[width=0.33\textwidth]{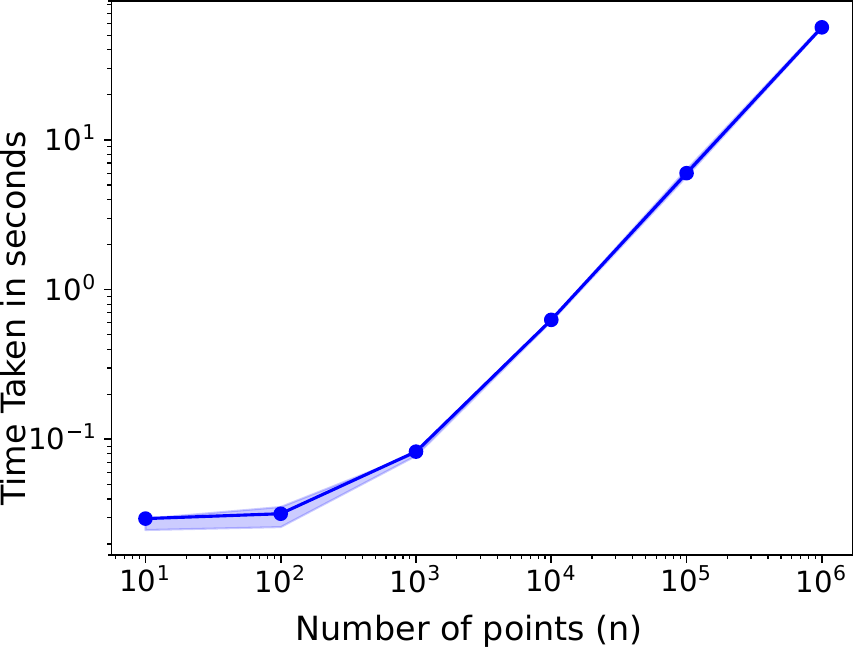}
  \vspace{-2mm}
  \caption{Runtime vs.\ number of rotations for \texttt{SO3\_SPMC}. The log-log slope is $\approx 1$ for $10^4$-$10^6$ points.}
  \label{fig:time_taken}
\end{figure}

Figure~\ref{fig:time_taken} plots wall-clock time vs.\ $n$. 
After a small warm-up region ($n\!\le\!10^3$), the log-log slope is approximately $1$ for $10^4$-$10^6$ points, empirically confirming near-linear complexity consistent with the theory.

\begin{figure}[ht]
  \centering
  \includegraphics[width=0.25\textwidth]{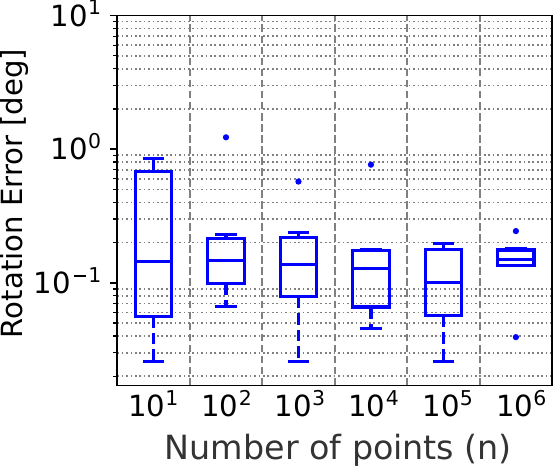}
  \vspace{-2mm}
  \caption{Rotational error vs.\ number of rotations for \texttt{SO3\_SPMC}.}
  \label{fig:rot_error_points}
\end{figure}

Figure~\ref{fig:rot_error_points} shows the rotational error vs.\ $n$. 
As $n$ increases, alignment improves: larger samples stabilize the TBV histograms and reduce sensitivity to outliers (the error distribution’s tail shrinks). 
At very small $n$ (e.g., $10$ points), the empirical means and histograms are more easily skewed, producing higher variance.

% \subsection{Discussion}
Overall, our numerical simulation results had accuracies 
of $0.068^\circ$ or less for 2000 points, while the accuracies for 10 points were much larger, between $0.08^\circ$-$0.15^\circ$. 
This may be due to the bin sized used, with $1^\circ$ bins.  For small numbers of data points, each point will be placed into a bin at a random location, and thus there will be some error between the original and rotated points.  Using additional bins would provide increased accuracy, and this would be computationally tractable for small numbers of points.  Alternatively, sub-pixel estimation schemes such as \cite{guizar2008efficient} could be used to improve the results.

\FloatBarrier

%\begin{thebibliography}{99}
%\end{thebibliography}

\bibliographystyle{ieee-tran}
\bibliography{citations}

@String(CVPR= {IEEE Conf. Comput. Vis. Pattern Recog.})

@String(ICCV= {Int. Conf. Comput. Vis.})

@String(CVPR  = {CVPR})

@String(ICCV  = {ICCV})

@inproceedings{sorgi2004normalized,
  title={Normalized cross-correlation for spherical images},
  author={Sorgi, Lorenzo and Daniilidis, Kostas},
  booktitle={Computer Vision-ECCV 2004: 8th European Conference on Computer Vision, Prague, Czech Republic, May 11-14, 2004. Proceedings, Part II 8},
  pages={542--553},
  year={2004},
  organization={Springer}
}

@article{cohen2018spherical,
  title={Spherical cnns},
  author={Cohen, Taco S and Geiger, Mario and K{\"o}hler, Jonas and Welling, Max},
  journal={arXiv preprint arXiv:1801.10130},
  year={2018}
}

@article{esteves2023scaling,
  title={Scaling spherical cnns},
  author={Esteves, Carlos and Slotine, Jean-Jacques and Makadia, Ameesh},
  journal={arXiv preprint arXiv:2306.05420},
  year={2023}
}

@article{makadia2006rotation,
  title={Rotation recovery from spherical images without correspondences},
  author={Makadia, Ameesh and Daniilidis, Kostas},
  journal={IEEE transactions on pattern analysis and machine intelligence},
  volume={28},
  number={7},
  pages={1170--1175},
  year={2006},
  publisher={IEEE}
}

@article{wahba1965least,
  title={A least squares estimate of satellite attitude},
  author={Wahba, Grace},
  journal={SIAM review},
  volume={7},
  number={3},
  pages={409--409},
  year={1965},
  publisher={Society for Industrial and Applied Mathematics}
}

@article{forbes2015linear,
  title={Linear-matrix-inequality-based solution to Wahba’s problem},
  author={Forbes, James Richard and de Ruiter, Anton HJ},
  journal={Journal of guidance, control, and dynamics},
  volume={38},
  number={1},
  pages={147--151},
  year={2015},
  publisher={American Institute of Aeronautics and Astronautics}
}

@article{horn1988closed,
  title={Closed-form solution of absolute orientation using orthonormal matrices},
  author={Horn, Berthold KP and Hilden, Hugh M and Negahdaripour, Shahriar},
  journal={Josa a},
  volume={5},
  number={7},
  pages={1127--1135},
  year={1988},
  publisher={Optica Publishing Group}
}

@article{khoshelham2016closed,
  title={Closed-form solutions for estimating a rigid motion from plane correspondences extracted from point clouds},
  author={Khoshelham, Kourosh},
  journal={ISPRS Journal of Photogrammetry and Remote Sensing},
  volume={114},
  pages={78--91},
  year={2016},
  publisher={Elsevier}
}

@article{markley1988attitude,
  title={Attitude determination using vector observations and the singular value decomposition},
  author={Markley, F Landis},
  journal={Journal of the Astronautical Sciences},
  volume={36},
  number={3},
  pages={245--258},
  year={1988}
}

@article{saunderson2015semidefinite,
  title={Semidefinite descriptions of the convex hull of rotation matrices},
  author={Saunderson, James and Parrilo, Pablo A and Willsky, Alan S},
  journal={SIAM Journal on Optimization},
  volume={25},
  number={3},
  pages={1314--1343},
  year={2015},
  publisher={SIAM}
}

@inproceedings{li20073d,
  title={The 3D-3D registration problem revisited},
  author={Li, Hongdong and Hartley, Richard},
  booktitle={2007 IEEE 11th international conference on computer vision},
  pages={1--8},
  year={2007},
  organization={IEEE}
}

@inproceedings{besl1992method,
  title={Method for registration of 3-D shapes},
  author={Besl, Paul J and McKay, Neil D},
  booktitle={Sensor fusion IV: control paradigms and data structures},
  volume={1611},
  pages={586--606},
  year={1992},
  organization={Spie}
}

@inproceedings{chetverikov2002trimmed,
  title={The trimmed iterative closest point algorithm},
  author={Chetverikov, Dmitry and Svirko, Dmitry and Stepanov, Dmitry and Krsek, Pavel},
  booktitle={2002 International Conference on Pattern Recognition},
  volume={3},
  pages={545--548},
  year={2002},
  organization={IEEE}
}

@article{horn1987closed,
  title={Closed-form solution of absolute orientation using unit quaternions},
  author={Horn, Berthold KP},
  journal={Josa a},
  volume={4},
  number={4},
  pages={629--642},
  year={1987},
  publisher={Optica Publishing Group}
}

@article{arun1987least,
  title={Least-squares fitting of two 3-D point sets},
  author={Arun, K Somani and Huang, Thomas S and Blostein, Steven D},
  journal={IEEE Transactions on pattern analysis and machine intelligence},
  volume={PAMI-9},
  number={5},
  pages={698--700},
  year={1987},
  publisher={IEEE}
}

@inproceedings{yang2019quaternion,
  title={A quaternion-based certifiably optimal solution to the Wahba problem with outliers},
  author={Yang, Heng and Carlone, Luca},
  booktitle={Proceedings of the IEEE/CVF International Conference on Computer Vision},
  pages={1665--1674},
  year={2019}
}

@inproceedings{LiaoCVPR2019,
  author={Liao, Shuai and Gavves, Efstratios and Snoek, Cees G. M.},
  booktitle={2019 IEEE/CVF Conference on Computer Vision and Pattern Recognition (CVPR)}, 
  title={Spherical Regression: Learning Viewpoints, Surface Normals and {3D} Rotations on {N}-Spheres}, 
  year={2019},
  volume={},
  number={},
  pages={9751-9759},
  doi={10.1109/CVPR.2019.00999}}

@article{henderson1996experiencing,
  title={Experiencing geometry: on plane and sphere},
  author={Henderson, David Wilson and Moura, Eduarda},
  journal={(No Title)},
  year={1996}
}

@book{flanders1963differential,
  title={Differential forms with applications to the physical sciences},
  author={Flanders, Harley},
  volume={11},
  year={1963},
  publisher={Courier Corporation}
}

@inproceedings{Thibaut2022,
       booktitle = {16{\`e}me Congr{\`e}s Fran{\c c}ais d?Acoustique},
           month = {April},
           title = {Spherical correlation as a similarity measure for 3D radiation patterns of musical instruments},
          author = {T. Carpentier and A. Einbond},
       publisher = {HAL Open Science},
            year = {2022},
             url = {https://openaccess.city.ac.uk/id/eprint/28202/}
}

@inproceedings{kazhdan2002harmonic,
  title={Harmonic 3D shape matching},
  author={Kazhdan, Michael and Funkhouser, Thomas},
  booktitle={ACM SIGGRAPH 2002 conference abstracts and applications},
  pages={191--191},
  year={2002}
}

@inproceedings{kazhdan2003rotation,
  title={Rotation invariant spherical harmonic representation of 3 d shape descriptors},
  author={Kazhdan, Michael and Funkhouser, Thomas and Rusinkiewicz, Szymon},
  booktitle={Symposium on geometry processing},
  volume={6},
  pages={156--164},
  year={2003}
}

@techreport{slabaugh1999computing,
  title={Computing Euler angles from a rotation matrix},
  author={Slabaugh, Gregory G},
  institution={CityUniv. London, London, U.K.},
  volume={6},
  number={2000},
  pages={39--63},
  year={1999}
}

@article{anik2022,
author={Sarker,Anik and Don-Roberts Emenonye and Kelliher,Aisling and Rikakis,Thanassis and Buehrer,R. M. and Asbeck, Alan T.},
year={2022},
title={Capturing Upper Body Kinematics and Localization with Low-Cost Sensors for Rehabilitation Applications},
journal={Sensors},
volume={22},
number={6},
pages={2300},
}

@article{markley2000quaternion,
  title={Quaternion attitude estimation using vector observations},
  author={Markley, F Landis and Mortari, Daniele},
  journal={The Journal of the Astronautical Sciences},
  volume={48},
  pages={359--380},
  year={2000},
  publisher={Springer}
}

@article{wu2017fast,
  title={Fast linear quaternion attitude estimator using vector observations},
  author={Wu, Jin and Zhou, Zebo and Gao, Bin and Li, Rui and Cheng, Yuhua and Fourati, Hassen},
  journal={IEEE Transactions on Automation Science and Engineering},
  volume={15},
  number={1},
  pages={307--319},
  year={2017},
  publisher={IEEE}
}

@techreport{keat1977analysis,
  title={Analysis of least-squares attitude determination routine DOAOP},
  author={Keat, J},
  year={1977},
  institution={Technical Report CSC/TM-77/6034, Comp. Sc. Corp}
}

@article{guizar2008efficient,
  title={Efficient subpixel image registration algorithms},
  author={Guizar-Sicairos, Manuel and Thurman, Samuel T and Fienup, James R},
  journal={Optics letters},
  volume={33},
  number={2},
  pages={156--158},
  year={2008},
  publisher={Optica Publishing Group}
}

@inproceedings{furrer2017evaluation,
  title={Evaluation of combined time-offset estimation and hand-eye calibration on robotic datasets},
  author={Furrer, Fadri and Fehr, Marius and Novkovic, Tonci and Sommer, Hannes and Gilitschenski, Igor and Siegwart, Roland},
  booktitle={Field and Service Robotics: Results of the 11th International Conference},
  pages={145--159},
  year={2017},
  organization={Springer}
}

@article{li2010simultaneous,
  title={Simultaneous robot-world and hand-eye calibration using dual-quaternions and {Kronecker} product},
  author={Li, Aiguo and Wang, Lin and Wu, Defeng},
  journal={Int. J. Phys. Sci},
  volume={5},
  number={10},
  pages={1530--1536},
  year={2010}
}

@article{wise2025certifably,
  title={A Certifably Correct Algorithm for Generalized Robot-World and Hand-Eye Calibration},
  author={Wise, Emmett and Kaveti, Pushyami and Chen, Qilong and Wang, Wenhao and Singh, Hanumant and Kelly, Jonathan and Rosen, David M and Giamou, Matthew},
  journal={arXiv preprint arXiv:2507.23045},
  year={2025}
}

@inproceedings{wang2022accurate,
  title={Accurate calibration of multi-perspective cameras from a generalization of the hand-eye constraint},
  author={Wang, Yifu and Jiang, Wenqing and Huang, Kun and Schwertfeger, S{\"o}ren and Kneip, Laurent},
  booktitle={2022 International Conference on Robotics and Automation (ICRA)},
  pages={1244--1250},
  year={2022},
  organization={IEEE}
}

@article{shah2013solving,
  title={Solving the robot-world/hand-eye calibration problem using the {Kronecker} product},
  author={Shah, Mili},
  journal={Journal of Mechanisms and Robotics},
  volume={5},
  number={3},
  pages={031007},
  year={2013},
  publisher={American Society of Mechanical Engineers}
}

@article{tabb2017solving,
  title={Solving the robot-world hand-eye (s) calibration problem with iterative methods},
  author={Tabb, Amy and Ahmad Yousef, Khalil M},
  journal={Machine Vision and Applications},
  volume={28},
  number={5},
  pages={569--590},
  year={2017},
  publisher={Springer}
}

@inproceedings{horn2023extrinsic,
  title={Extrinsic infrastructure calibration using the hand-eye robot-world formulation},
  author={Horn, Markus and Wodtko, Thomas and Buchholz, Michael and Dietmayer, Klaus},
  booktitle={2023 IEEE Intelligent Vehicles Symposium (IV)},
  pages={1--8},
  year={2023},
  organization={IEEE}
}

@article{evangelista2023graph,
  title={A graph-based optimization framework for hand-eye calibration for multi-camera setups},
  author={Evangelista, Daniele and Olivastri, Emilio and Allegro, Davide and Menegatti, Emanuele and Pretto, Alberto},
  journal={arXiv preprint arXiv:2303.04747},
  year={2023}
}

@article{ulrich2023uncertainty,
  title={Uncertainty-aware hand--eye calibration},
  author={Ulrich, Markus and Hillemann, Markus},
  journal={IEEE Transactions on Robotics},
  volume={40},
  pages={573--591},
  year={2023},
  publisher={IEEE}
}

@article{ransac_original,
author = {Fischler, Martin A. and Bolles, Robert C.},
title = {Random sample consensus: a paradigm for model fitting with applications to image analysis and automated cartography},
year = {1981},
issue_date = {June 1981},
publisher = {Association for Computing Machinery},
address = {New York, NY, USA},
volume = {24},
number = {6},
issn = {0001-0782},
url = {https://doi.org/10.1145/358669.358692},
doi = {10.1145/358669.358692},
journal = {Commun. ACM},
month = jun,
pages = {381–395},
numpages = {15}
}

@inproceedings{schmidt2003robust,
  title={Robust hand--eye calibration of an endoscopic surgery robot using dual quaternions},
  author={Schmidt, Jochen and Vogt, Florian and Niemann, Heinrich},
  booktitle={Joint Pattern Recognition Symposium},
  pages={548--556},
  year={2003},
  organization={Springer}
}

@article{burri2016euroc,
  title={The {EuRoC} Micro Aerial Vehicle Datasets},
  author={Burri, Michael and Nikolic, Janosch and Gohl, Pascal and Schneider, Thomas and Rehder, Joern and Omari, Sammy and Achtelik, Markus W and Siegwart, Roland},
  journal={The International Journal of Robotics Research},
  volume={35},
  number={10},
  pages={1157--1163},
  year={2016},
  publisher={SAGE Publications Sage UK: London, England}
}

@inproceedings{sarker2025correspondence,
  title     = {Correspondence-Free Fast and Robust Spherical Point Pattern Registration},
  author    = {Sarker, Anik and Asbeck, Alan T.},
  booktitle = {Proceedings of the IEEE/CVF International Conference on Computer Vision (ICCV)},
  year      = {2025},
  pages     = {28156--28166}
}

@inproceedings{Seel2012JointAxis,
  author    = {Thomas Seel and Thomas Schauer and J{\"o}rg Raisch},
  title     = {Joint axis and position estimation from inertial measurement data by exploiting kinematic constraints},
  booktitle = {Proc. IEEE Int. Conf. on Control Applications (CCA)},
  year      = {2012},
  pages     = {45--49},
  doi       = {10.1109/CCA.2012.6402433}
}

@article{Seel2014IMUAngles,
  author  = {Thomas Seel and Jonas Raisch and Thomas Schauer},
  title   = {{IMU}-Based Joint Angle Measurement for Gait Analysis},
  journal = {Sensors},
  year    = {2014},
  volume  = {14},
  number  = {4},
  pages   = {6891--6909},
  doi     = {10.3390/s140406891}
}

@article{Ekdahl2023S2S,
  author  = {M. Ekdahl and K. D. Dahl and J. R. C. Hewett and others},
  title   = {Inertial Measurement Unit Sensor-to-Segment Calibration Comparison for Sport-Specific Motion Analysis},
  journal = {Sensors},
  year    = {2023},
  volume  = {23},
  number  = {18},
  pages   = {7987},
  doi     = {10.3390/s23187987}
}

@article{Olsson2020PnP,
  author  = {Fredrik Olsson and Manon Kok and Thomas Seel and Kristoffer Halvorsen},
  title   = {Robust Plug-and-Play Joint Axis Estimation Using Inertial Sensors},
  journal = {Sensors},
  year    = {2020},
  volume  = {20},
  number  = {12},
  pages   = {3534},
  doi     = {10.3390/s20123534}
}

@article{DiRaimondo2022S2SReview,
  author  = {Giuseppe Di Raimondo and Federico Magistri and Alessandra Pedrocchi and others},
  title   = {A Scoping Review of Sensor-to-Segment Calibration for Wearable Inertial Sensors},
  journal = {Sensors},
  year    = {2022},
  volume  = {22},
  number  = {24},
  pages   = {9837},
  doi     = {10.3390/s22249837}
}

@article{Liu2014MagCal,
  author  = {Yuxin Liu and Yuxin Zhang and Min Tan and Kaimin Xu and Yihua Li and Xin Wang},
  title   = {Novel Calibration Algorithm for a Three-Axis Strapdown Magnetometer},
  journal = {Sensors},
  year    = {2014},
  volume  = {14},
  number  = {5},
  pages   = {8485--8504},
  doi     = {10.3390/s140508485}
}

@article{Pang2013MagCal,
  author  = {Chong Pang and Lei Chen and Yanping Liu and Shuo Wang},
  title   = {Calibration and compensation of three-axis magnetometer errors by using a nonlinear ellipsoid fitting algorithm},
  journal = {Journal of Magnetism and Magnetic Materials},
  year    = {2013},
  volume  = {346},
  pages   = {177--182},
  doi     = {10.1016/j.jmmm.2013.06.047}
}

@article{Wu2020MagCal,
  author  = {Hua Wu and Shengjun Liu and Zhen Wei and Zhiyong Zhang and Ying Sun},
  title   = {An Improved Magnetometer Calibration and Compensation Method Based on {Levenberg--Marquardt} Algorithm},
  journal = {Measurement and Control},
  year    = {2020},
  volume  = {53},
  number  = {9-10},
  pages   = {1312--1321},
  doi     = {10.1177/0020294019890627}
}

@inproceedings{Cui2018MagCal,
  author    = {Xiaoliang Cui and Anqin Zhang and Changping Qi and Shiyong Zhang},
  title     = {An Accurate Calibration Method for Magnetometers Based on Ellipsoid Fitting},
  booktitle = {Proc. IEEE/ION Position, Location and Navigation Symposium (PLANS)},
  year      = {2018},
  pages     = {1010--1015},
  doi       = {10.1109/PLANS.2018.8373519}
}

@article{ShiuAhmad1989AXXB,
  author  = {Ying-Shen Shiu and Shaheen Ahmad},
  title   = {Calibration of Wrist-Mounted Robotic Sensors by Solving Homogeneous Transform Equations of the Form {AX}= {XB}},
  journal = {IEEE Transactions on Robotics and Automation},
  year    = {1989},
  volume  = {5},
  number  = {1},
  pages   = {16--29},
  doi     = {10.1109/70.34770}
}

@article{Li2010RWHE,
  author  = {Aiguo Li and Linjun Wang and Danwei Wang},
  title   = {Simultaneous robot-world and hand-eye calibration using dual-quaternions and {Kronecker} product},
  journal = {The International Journal of Robotics Research},
  year    = {2010},
  volume  = {29},
  number  = {5},
  pages   = {635--653},
  doi     = {10.1177/0278364909356602}
}

@article{Shah2013Kronecker,
  author  = {Mili Shah and Sachin V. Naik and Jin G. L. and R. M. Voyles},
  title   = {Hand-eye calibration using {Kronecker} products},
  journal = {Mechanism and Machine Theory},
  year    = {2013},
  volume  = {69},
  pages   = {54--66},
  doi     = {10.1016/j.mechmachtheory.2013.05.006}
}

@article{Tabb2017RWHE,
  author  = {Amy Tabb and Khalil M. A. Yousef},
  title   = {Solving the Robot-World/Hand-Eye(s) Calibration Problem with Iterative Methods},
  journal = {Machine Vision and Applications},
  year    = {2017},
  volume  = {28},
  number  = {5–6},
  pages   = {569--590},
  doi     = {10.1007/s00138-017-0841-7}
}

@incollection{Furrer2018TimeOffset,
  author    = {Fadri Furrer and Marius Fehr and Tonci Novkovic and Hannes Sommer and Igor Gilitschenski and Roland Siegwart},
  title     = {Evaluation of Combined Time-Offset Estimation and Hand-Eye Calibration on Robotic Datasets},
  booktitle = {Field and Service Robotics},
  series    = {Springer Proceedings in Advanced Robotics},
  volume    = {5},
  year      = {2018},
  pages     = {145--159},
  doi       = {10.1007/978-3-319-67361-5_10},
  publisher = {Springer}
}

@inproceedings{Mourikis2007MSCKF,
  author    = {Anastasios I. Mourikis and Stergios I. Roumeliotis},
  title     = {A Multi-State Constraint Kalman Filter for Vision-aided Inertial Navigation},
  booktitle = {Proc. IEEE Int. Conf. on Robotics and Automation (ICRA)},
  year      = {2007},
  pages     = {3565--3572},
  doi       = {10.1109/ROBOT.2007.364024}
}

@article{Qin2018VINSMono,
  author  = {Tong Qin and Peiliang Li and Shaojie Shen},
  title   = {VINS-Mono: A Robust and Versatile Monocular Visual-Inertial State Estimator},
  journal = {IEEE Transactions on Robotics},
  year    = {2018},
  volume  = {34},
  number  = {4},
  pages   = {1004--1020},
  doi     = {10.1109/TRO.2018.2853729}
}

@article{Leutenegger2015OKVIS,
  author  = {Stefan Leutenegger and Simon Lynen and Michael Bosse and Roland Siegwart and Paul Furgale},
  title   = {Keyframe-Based Visual–Inertial Odometry Using Nonlinear Optimization},
  journal = {The International Journal of Robotics Research},
  year    = {2015},
  volume  = {34},
  number  = {3},
  pages   = {314--334},
  doi     = {10.1177/0278364914554813}
}

@article{Scaramuzza2019VIO,
  author  = {Davide Scaramuzza and Zhen Zhang},
  title   = {Visual–Inertial Odometry},
  journal = {Encyclopedia of Robotics},
  year    = {2019},
  pages   = {1--8},
  doi     = {10.1007/978-3-642-41610-1_60-1},
  publisher = {Springer}
}

%%% Moved Appendices to Supplementary Material, that is what you are supposed to do for this journal

\end{document}